\documentclass[runningheads]{llncs}
\usepackage{graphicx}
\usepackage[outdir=./]{epstopdf}
\usepackage{subfigure}
\usepackage{amsmath,amssymb} 
\usepackage{color}
\usepackage[width=122mm,left=12mm,paperwidth=146mm,height=193mm,top=12mm,paperheight=217mm]{geometry}

\def\v#1{\mathbf{#1}}
\def\m#1{\mathsf{#1}}

\def\tr#1{{#1}^\mathsf{T}}

\begin{document}
\pagestyle{headings}
\mainmatter

\title{Similarity Registration Problems for 2D/3D Ultrasound Calibration} 

\titlerunning{Similarity Registration Problems for 2D/3D US Calibration}

\authorrunning{Francisco Vasconcelos \and Donald Peebles \and Sebastien Ourselin \and Danail Stoyanov}

\author{F. Vasconcelos \and D. Peebles \and S. Ourselin \and D. Stoyanov}


\institute{University College London, United Kingdom\\
	\email{ \{v.vasconcelos,d.peebles,s.ourselin,danail.stoyanov\}@ucl.ac.uk}
}

\maketitle

\begin{abstract}
We propose a minimal solution for the similarity registration (rigid pose and scale) between two sets of 3D lines, and also between a set of co-planar points and a set of 3D lines. The first problem is solved up to 8 discrete solutions with a minimum of 2 line-line correspondences, while the second is solved up to 4 discrete solutions using 4 point-line correspondences. We use these algorithms to perform the extrinsic calibration between a pose tracking sensor and a 2D/3D ultrasound (US) curvilinear probe using a tracked needle as calibration target. The needle is tracked as a 3D line, and is scanned by the ultrasound as either a 3D line (3D US) or as a 2D point (2D US). Since the scale factor that converts US scan units to metric coordinates is unknown, the calibration is formulated as a similarity registration problem. We present results with both synthetic and real data and show that the minimum solutions outperform the correspondent non-minimal linear formulations.

\end{abstract}

\section{Introduction}

Ultrasound (US) is a low-cost and real-time medical imaging technique in minimally invasive surgery and in percutaneous procedures. It observes information under the surface, so it is used to locate invisible details about vessels, nerves, or tumours. By tracking the pose of a 2D ultrasound probe (2D US) we can render 3D reconstructions from a collection of 2D slices \cite{Khamene2005}, while a tracked 3D probe (3D US) is able to build large and detailed 3D models from a set of 3D scans \cite{Brattain2011}. Both 2D US and 3D US can also be used to guide other tracked medical instruments, such as biopsy needles \cite{Stoll2012}, and fuse data with other imaging modalities such as endoscopes.

Freehand 3D ultrasound generally refers to the extrinsic calibration between a hand-held US probe and a pose tracking device. This calibration aims at determining the rigid transformation between the US scan and the tracked marker as well as the scale factor that converts the US scan to metric coordinates, i. e. a similarity transformation. This is usually achieved by scanning a known calibration object (phantom) immersed in either water or a tissue mimicking gel. Since the speed of sound in water is different than in tissue, sometimes an alcoholic solution is used to obtain a more realistic US scale. A multitude of calibration phantoms with different shapes have been proposed in the literature \cite{Mercier2005}, including intersecting wires \cite{Chen2009}, a single plane \cite{Prager1998,Najafi2015}, a stylus \cite{Muratore2001,Khamene2005,Hsu2008}, and 3D printed objects \cite{Najafi2014}. Although these methods focus on 2D US calibration, some extensions to 3D US using similar phantoms have been proposed as well \cite{Bergmeir2009,Hummel2013}. 

In this paper we focus on using a tracked needle as the calibration phantom. Our main motivation is towards assisted guidance and motion analysis in fetal interventions that require the extraction of \textit{in utero} samples with a biopsy needle. It thus becomes a practical solution to use the same needle as a calibration object, avoiding the need to introduce new objects in the operating room and the additional burden of their sterilization. The tracked needle is detected by the pose tracking system as a 3D line, and it is scanned either as a line (3D US) or as a point (2D US). By scanning the needle under different poses, we formulate the 3D US calibration as the similarity registration between two sets of 3D lines and the 2D US calibration as the similarity registration between co-planar 3D points and 3D lines.

In this paper we propose a minimal solver to the similarity registration between two sets of 3D lines. We will also show that the registration between co-planar 3D points and 3D lines is a sub-problem of the same formulation and therefore the same minimal solver can be applied. Additionally, we show that this minimal solution can be easily generalised to the registration of any combination of plane, line, and point correspondences. We also present an alternative simplified minimal solver to the similarity registration between a set of co-planar points and a set of 3D lines. We apply the minimal solutions to the calibration of a 2D US and a 3D US with a pose tracking sensor and perform validation with both synthetic and real data.

\begin{figure}[t]
\centering
\subfigure[]{\includegraphics[width=0.315\textwidth]{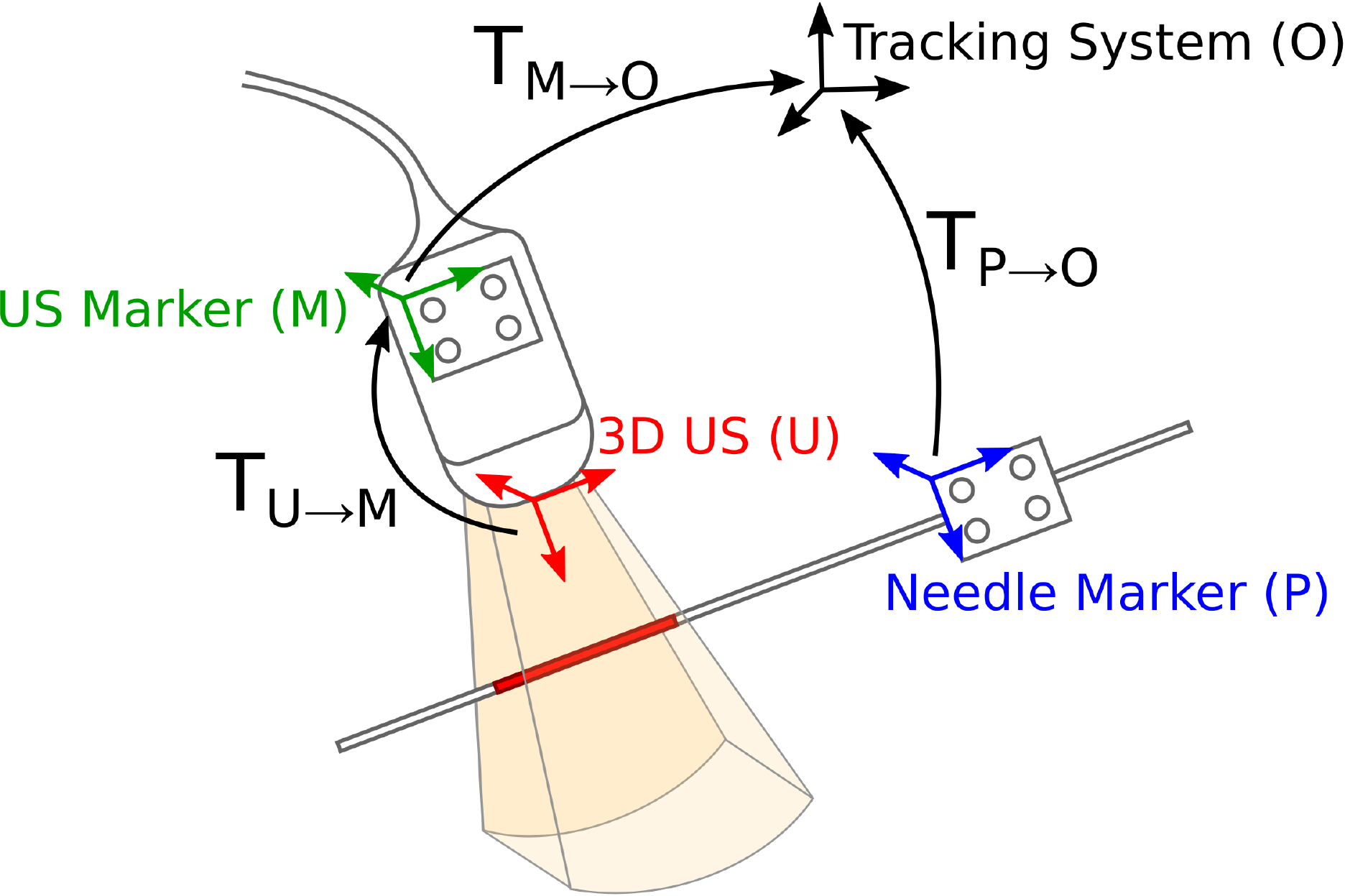}
}
\subfigure[]{\includegraphics[width=0.27\textwidth]{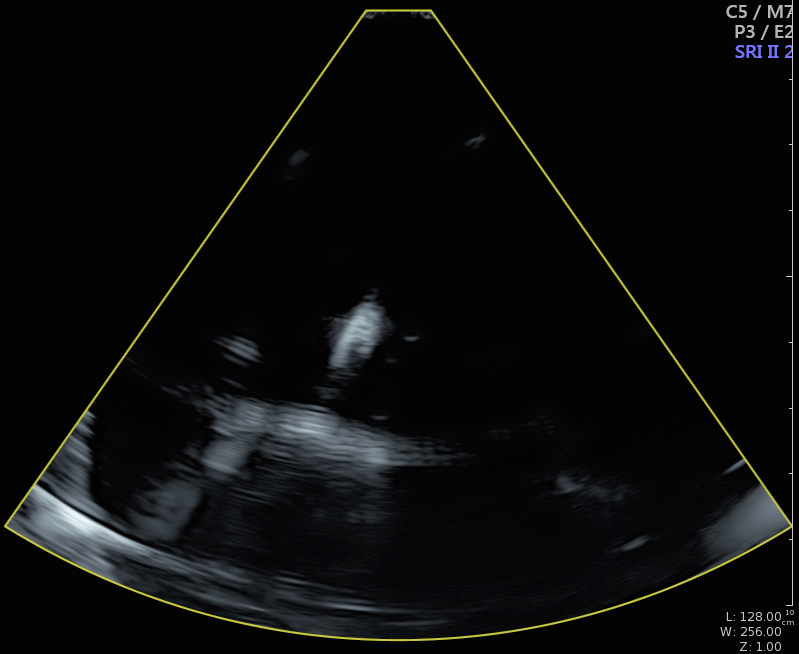}
}
\subfigure[]{\includegraphics[width=0.28\textwidth]{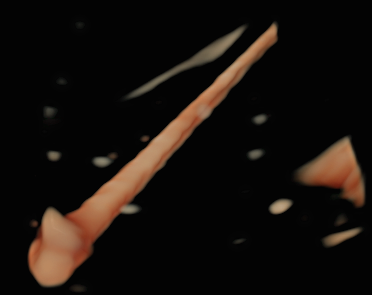}
}
\label{fig:NeedleScan}
\caption{(a) Scanning a tracked needle with a US probe; (b) a 2D US probe detects a cross section of the needle; (c) a 3D US detects a line segment.}
\end{figure}

\section{Related Work}

Freehand US calibration using a tracked linear target was proposed in \cite{Khamene2005}. However, this method is initialized with a non-minimal linear solution and is only meant for calibration of 2D US probes. Furthermore, it assumes that the US probe produces an anisotropic image, i. e. it has different scaling factors along the $x$ and $y$ axes of the image. An alternative method \cite{Vasconcelos2016} extends this calibration procedure to 3D US and shows that assuming an isotropic model (single scale factor) produces better calibration results for curvilinear shaped probes. In this paper we assume that the US scans are isotropic. In some contexts, it is possible to assume that the scale factor is known, and the calibration problem becomes the Euclidean registration between the US probe and the phantom target. In the 2D US case this problem becomes equivalent to the extrinsic calibration between a camera and a laser \textit{rangefinder} \cite{Zhang2004}. In the 3D US case, with the appropriate phantom (e. g. 3 known 3D points) the absolute pose of the probe can be recovered in each calibration scan, and thus it can be formulated as the standard hand-eye problem \cite{Horaud1995,Thompson2016}. In this paper, however, we consider that the scale is always unknown.

Estimating the similarity transformation (rigid pose and scale) between two coordinate frames gained recent attention due to its application in the registration of different Structure-from-Motion (SfM) sequences. If the same scene is recovered in two different monocular SfM runs, the scale of each reconstruction can be arbitrarily different. Therefore, to produce extended and more detailed 3D maps from independent SfM runs both the rigid pose and the scale must be recovered. If correspondences between SfM sequences are not available, one can use an extension of the ICP algorithm \cite{Zhang1994} to handle unkown scale \cite{Du2010}. If 2D-3D point correspondences are available, this is called the generalised pose and scale problem \cite{Ventura2014,Sweeney2014}, and is solved by extending the $PnP$ formulation \cite{Haralick1991,Quan1999,Lepetit2009} to handle the alignment of image rays from multiple view points. A closely related contribution estimates a similarity transformation from pairwise point correspondences between two generalised cameras \cite{Sweeney2015}. 

In the case of 3D US calibration, we are interested in the similarity registration between two sets of 3D lines. Different algorithms have been proposed to the euclidean registration between sets of lines \cite{Zhang1991,Kamgar2004}. One possible approach to solve the similarity registration problem would be to first estimate the unknown scale factor independently, e. g. by computing the ratio of orthogonal distances between all pairs of lines in both sets, and then use any of the previously mentioned euclidean registration algorithms. We found that this approach is extremely unstable with noisy measurements and thus we focus on the joint estimation of all similarity parameters. Non-minimal linear algorithms and non-linear refinement methods have been proposed to solve the registration of two sets of 3D lines for different non-rigid configurations, including the similarity transformation \cite{Bartoli2001,Anonymous}. However, and to the best of our knowledge, a minimal closed-form solution for the similarity registration of two sets of lines have not been proposed in the literature.  

The 2D US calibration problem is the similarity registration between a set of 3D lines and a set of co-planar points. This is a particular case of the pose and scale problem \cite{Ventura2014} when the 3D points are co-planar, and therefore this method could be adapted to solve this problem. However, the co-planarity of points introduces further simplifications, and as we will show in this paper, this problem can be minimally solved with a much more compact set of equations.

Our strategy to solve both registration problems is to convert them to an equivalent registration between a set of 3D points and and a set of 3D planes. Although this strategy has been described in the context of euclidean registration \cite{Ramalingam2010}, it is also valid for non-rigid registration.

Minimal solutions are a well established topic in computer vision literature \cite{Nister2004,Stewenius2005,Stewenius2005b,Kukelova2008,Kukelova2012}. In most cases they require solving a system of polynomial equations, which can be achieved using Grobner basis methods \cite{Stewenius2005,Byrod2009,Kukelova2012}. Although these methods provide a general framework to build numeric polynomial solvers, they require a certain amount of symbolic manipulation that often requires a case-by-case analysis. To address this issue an automatic generator of polynomial solvers have been proposed \cite{Kukelova2008}. In this paper we develop minimum solutions using the action matrix method as presented in \cite{Byrod2009}.



\section{2D/3D US Model}

\begin{figure}[tp]
\centering
\subfigure[]{\includegraphics[width=0.235\textwidth]{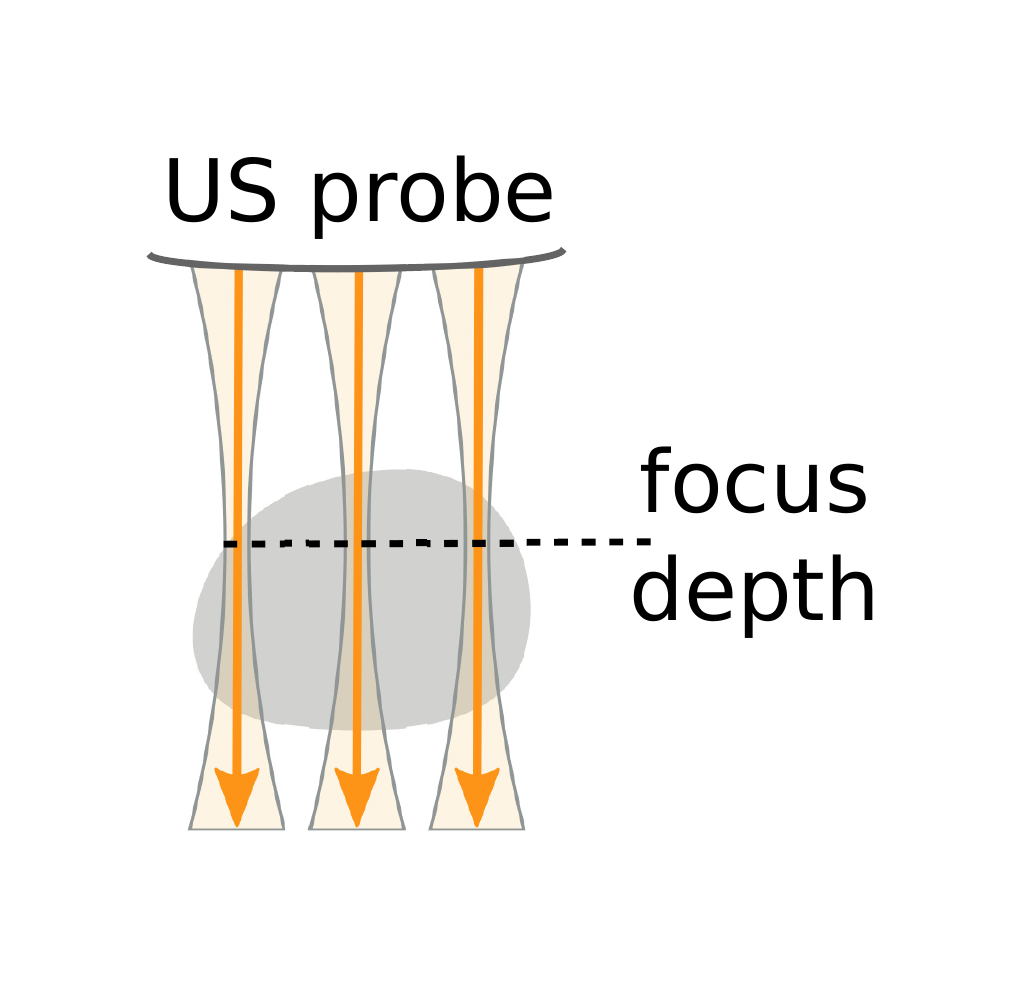}
\label{fig:USbeam}
}
\subfigure[]{\includegraphics[width=0.235\textwidth]{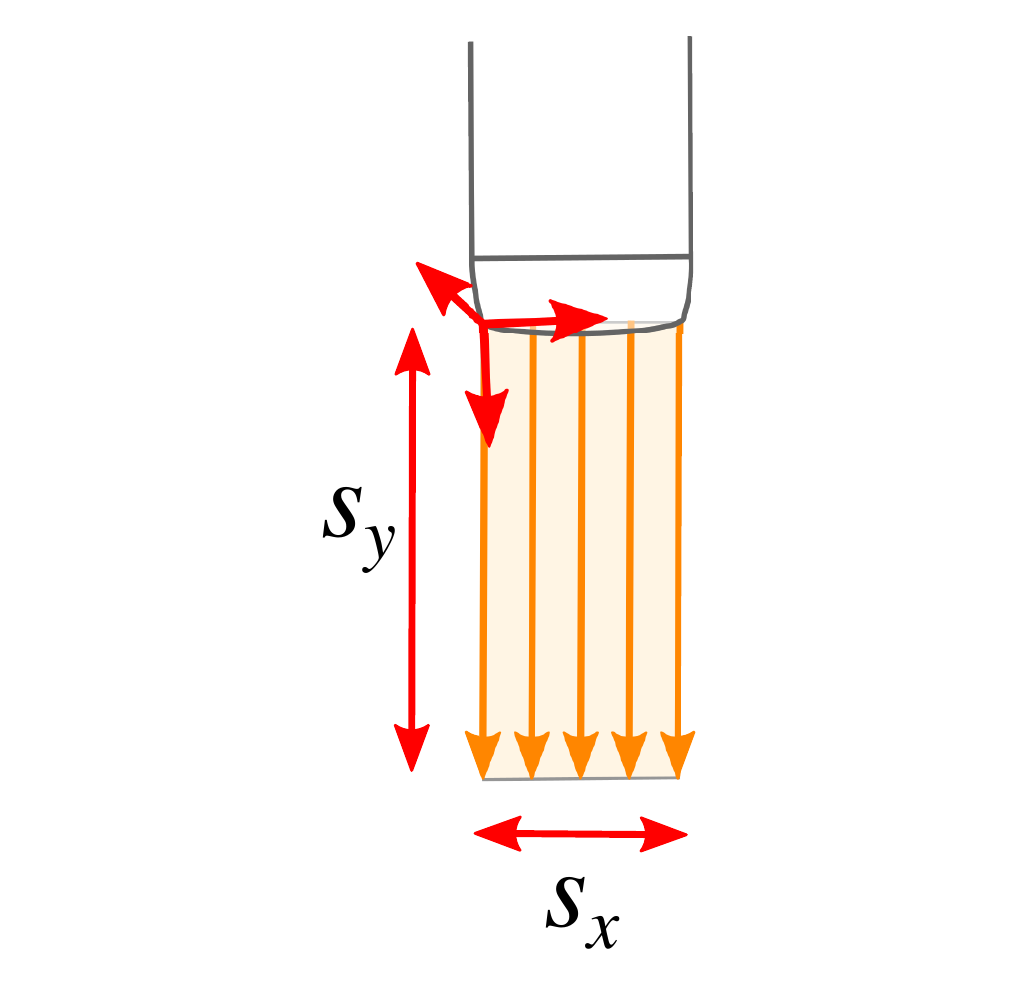}
\label{fig:model2DUSlin}
}
\subfigure[]{\includegraphics[width=0.235\textwidth]{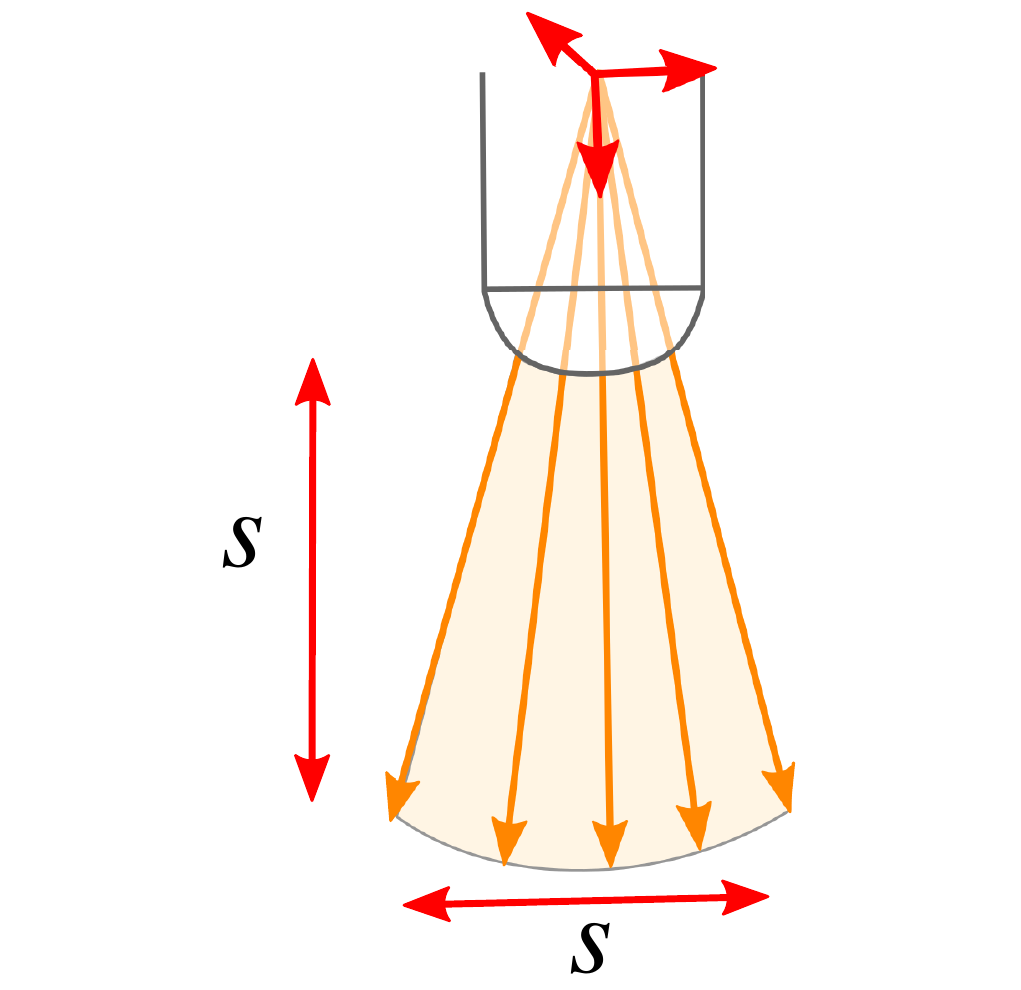}
\label{fig:model2DUScurv}
}
\subfigure[]{\includegraphics[width=0.235\textwidth]{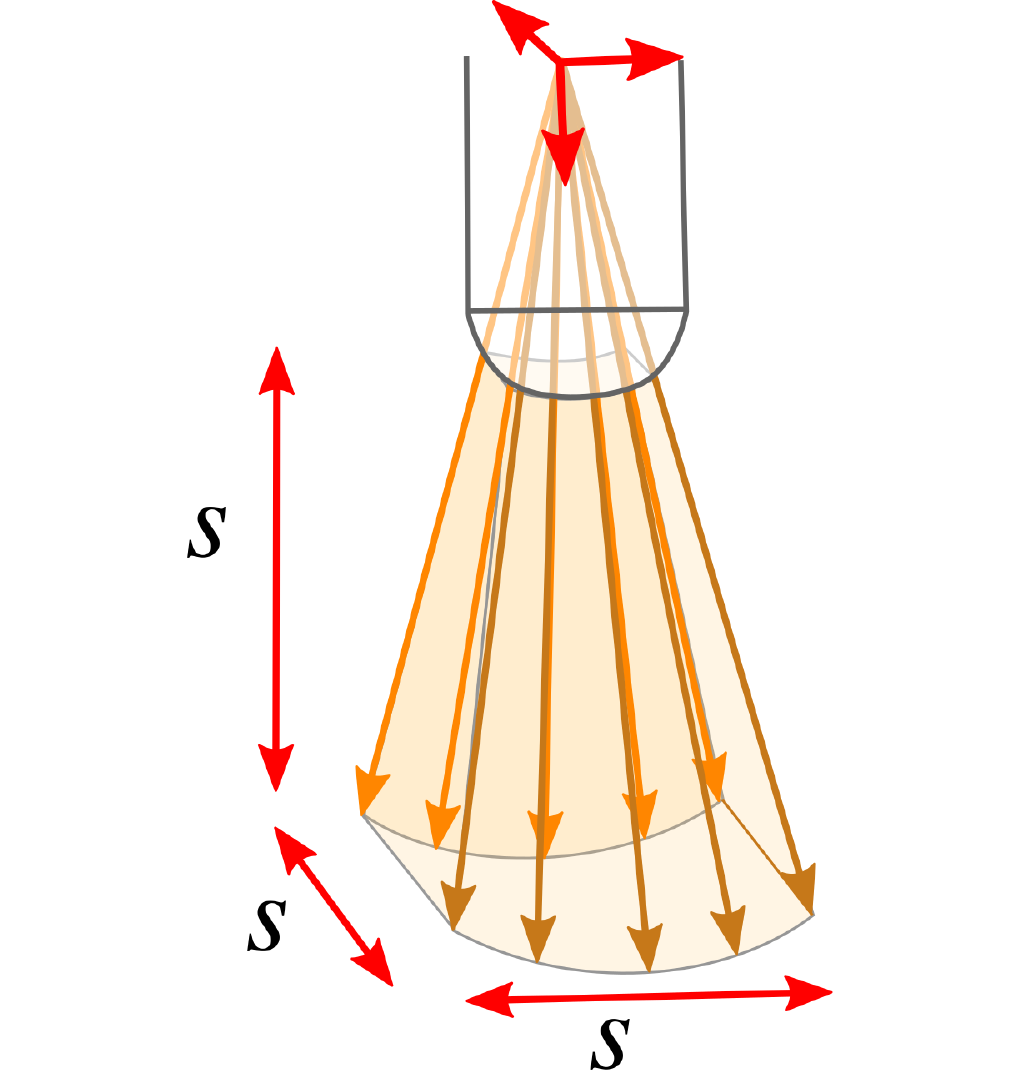}
\label{fig:model3DUScurv}
}
\caption{US probe models: (a) US emitted beams have a varying width. If we assume that the scanned region is focused, beams are approximated by a straight line. (b) Linear 2D US (c) Curvilinear 2D US (d) Curvilinear 3D US.}
\end{figure}

US probes emit a set of acoustic beams that are partially reflected whenever they cross a medium interface with a change in acoustic impedance. The time response of the echo reflections enables the formation of a spatial grayscale map representing the different acoustic impedances within the US scanning field of view. Note that US beams have a varying width (Fig. \ref{fig:USbeam}) which might induce undesired out of focus distortions. In an analogous way to most camera calibration models, we assume that the scanned region is always focused and thus each scene point reflects a beam along a single straight line. 

US image formation depends on the probe construction. Linear 2D US probes (Fig. \ref{fig:model2DUSlin}) emit parallel beams and thus there are two scale factors involved: $s_{y}$ depends on the speed of sound in the propagation medium, while $s_{x}$ is a fixed parameter that depends on the physical distance between beam emitters. These probes usually operate with high frequency acoustic signals (4 -- 16 MHz) and are used for short range scans (e. g. musculoskeletal imaging). The calibration of these probes cannot be represented with a similarity transformation and we discard it from further analysis in this paper. Curvilinear 2D US probes (Fig. \ref{fig:model2DUScurv}) emit beams in radial directions that intersect in a single point, forming a planar bundle of lines. In this case, the speed of sound in the propagation medium affects the scan scale isotropically. The curvilinear 3D US (Fig. \ref{fig:model3DUScurv}) is a generalization of the curvilinear 2D US, emitting a 3D bundle of beams. Curvilinear probes usually operate with lower frequency signals (2 -- 8 MHz) and are more suitable to long range scans (e. g. obstetrics, cardiac imaging).

\section{Problem Statement}

\begin{figure}[t]
\centering
\includegraphics[width=0.6\textwidth]{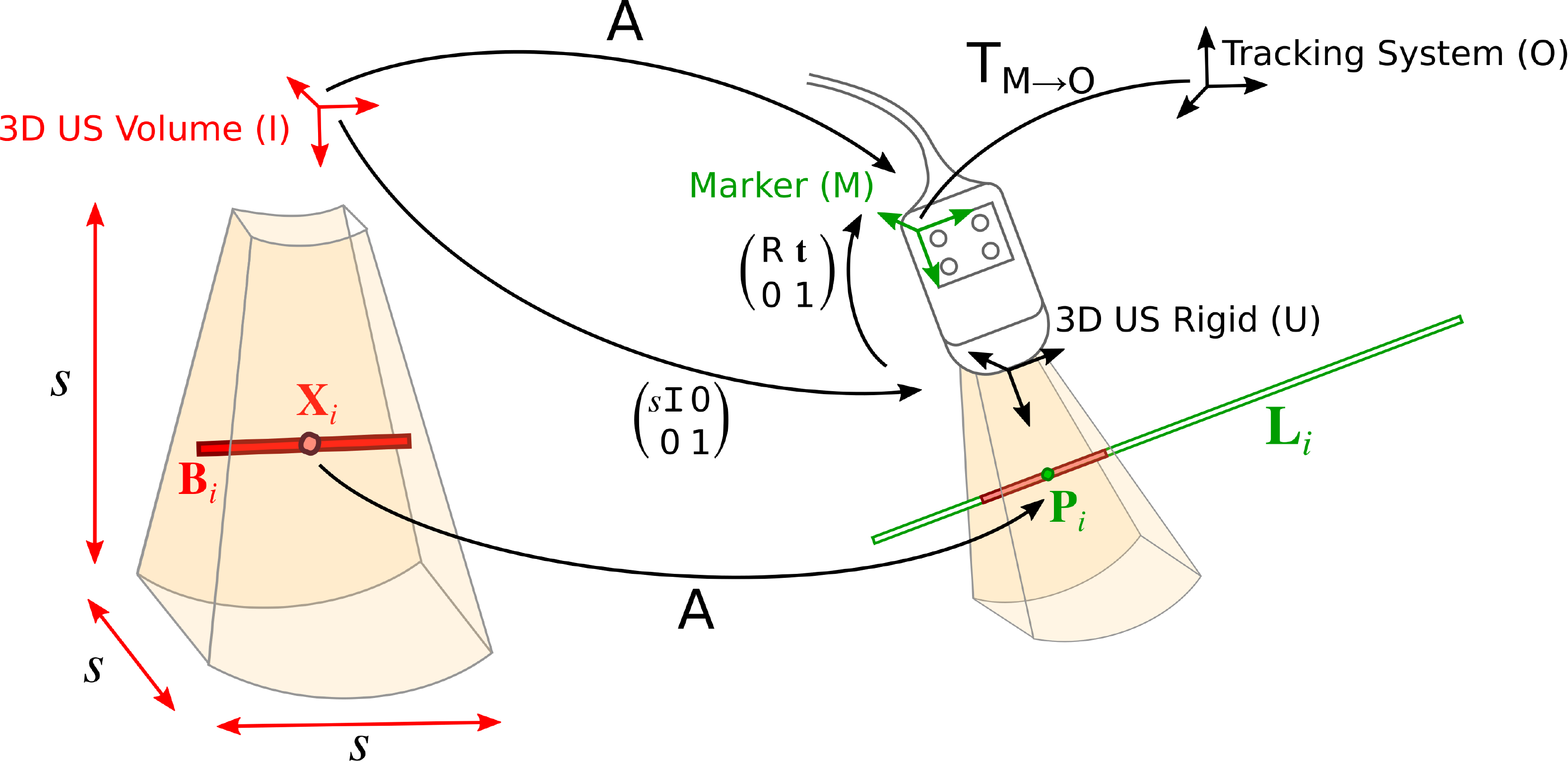}
\label{fig:3DCalibSetup}
\caption{3D US calibration problem. The similarity transformation $\m A$ maps points $\v X_{i}$ in the 3D US volume (red) to points $\v P_{i}$ in the marker reference frame (green). $\m A$ can be decomposed as a uniform scaling transformation followed by a rigid transformation.}
\end{figure}

Consider a hand-held curvilinear 3D US probe whose pose is tracked by a rigidly attached marker (Fig. 3). In each frame the tracking system determines the transformation $\m T_{\m M \rightarrow \m O}$ from the marker coordinate system ($\m M$) to a fixed frame $\m O$. The freehand 3D US calibration consists in determining the unknown similarity transformation $\m A$ that maps 3D points $\v X_{i}$ in the US volume to 3D points $\v P_{i}$ represented in $\m M$.
\begin{equation}
\label{eq:3DUSpointmapping}
\v P_{i} = \m A \v X_{i}
\end{equation}
The similarity $\m A$ is defined by a rotation $\m R$, a translation $\v t$ and a scale factor $s$ that converts the 3D US volume to metric coordinates, and is represented as
\begin{equation}
\m A = \begin{pmatrix}
\m S & \v t \\ 0 & 1
\end{pmatrix}
\end{equation}
where $\m S = s \m R$ is a scaled rotation matrix such that
\begin{equation}
\label{eq:similarityNonLin}
\tr{\m S} \m S = \m S \tr{\m S} = \begin{pmatrix}
s^{2} & 0 & 0 \\ 
0 & s^{2} & 0 \\
0 & 0 & s^{2}
\end{pmatrix}
\end{equation}

The calibration procedure consists in capturing a tracked needle in the 3D US volume under different poses. The needle is previously calibrated by determining its two endpoints in the reference frame $\m M$ and then it is represented in each acquisition as a 3D line $\v L_{i}$. The needle is also detected as a 3D line $\v B_{i}$ in the 3D US volume. The calibration problem is thus formulated as the 3D similarity registration between two sets of lines (Fig. \ref{fig:lineline3D}). 

\section{3D US Calibration Solution}

In this section we derive a minimal solution for the calibration of a 3D US probe. We start by re-stating it as the similarity registration between 3D points and 3D planes, and then derive a linear and a minimal solution for this problem. The calibration of a 2D US is presented in section \ref{2DUS} as a particular case of the 3D US problem.

\subsection{3D US Calibration as Point-Plane Registration}

\begin{figure}[t]
\centering
\subfigure[]{
\includegraphics[width=0.47\textwidth]{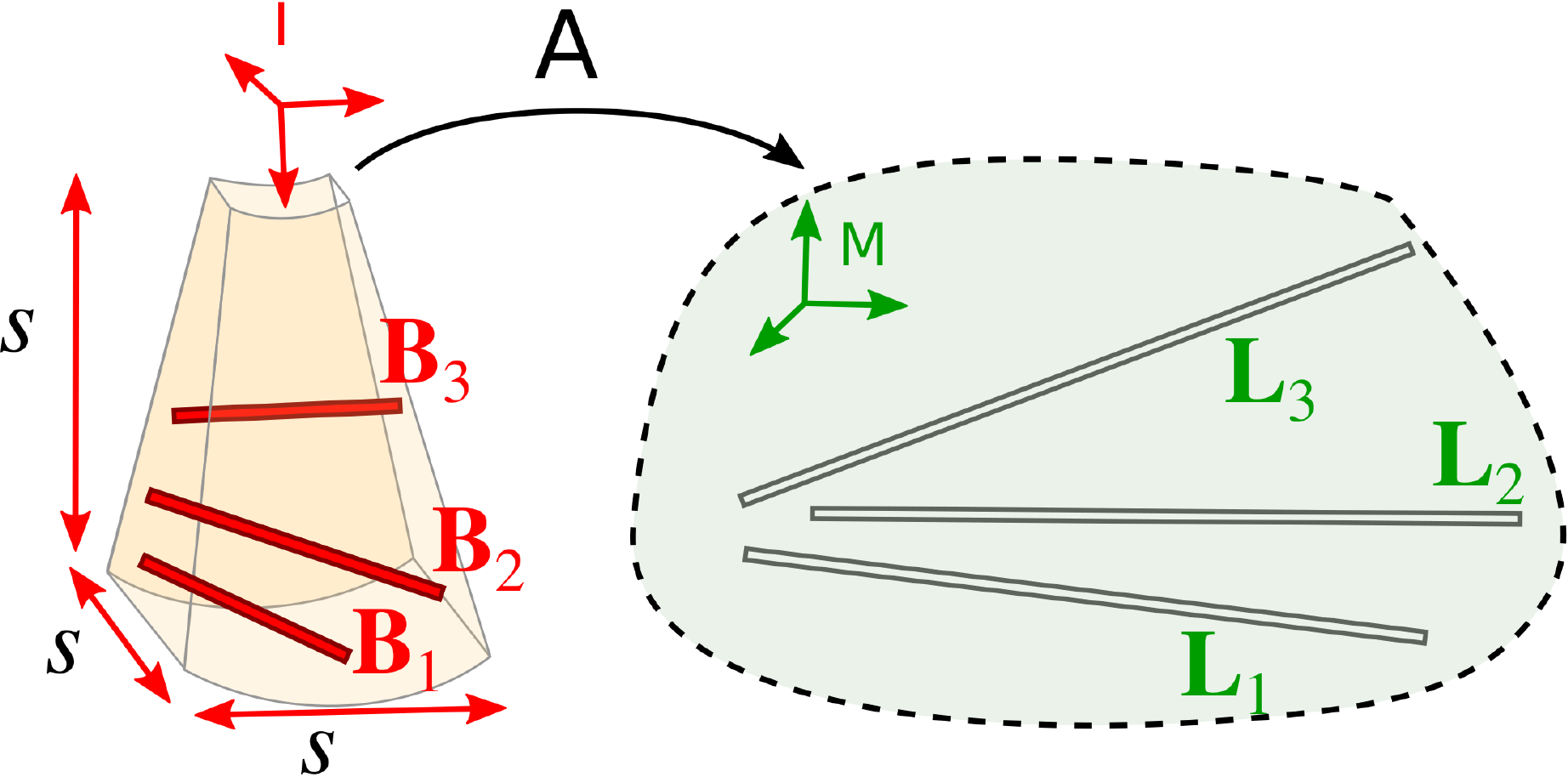}
\label{fig:lineline3D}
}
\subfigure[]{
\includegraphics[width=0.47\textwidth]{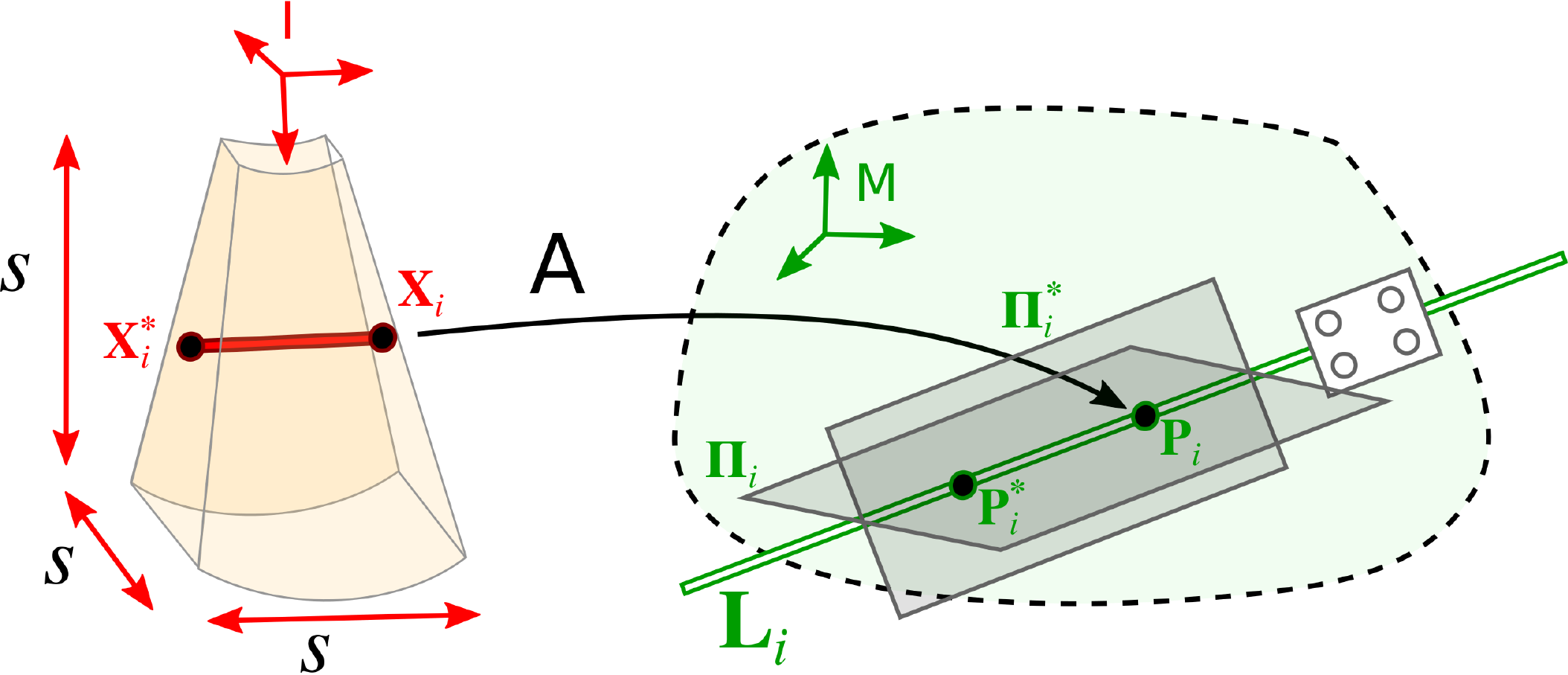}
\label{fig:ramalingam3D}
}
\caption{(a) The 3D US calibration is formulated as the similarity registration between lines $\v B_{i}$ and lines $\v L_{i}$; (b) Each line $\v L_{i}$ can be re-defined as two intersecting planes $\v \Pi_{i}$, $\v \Pi^{*}_{i}$, while each line $\v B_{i}$ can be redefined as two points $\v X_{i}$, $\v X^{*}_{i}$.}
\end{figure}

Ramalingam et. al. showed that any 3D registration problem involving 3D planes, lines and/or points can be re-stated as the registration between 3D planes and 3D points \cite{Ramalingam2010}. In our calibration problem this can be achieved by defining each needle line $\v L_{i}$ as two intersecting planes $\v \Pi_{i}$, $\v \Pi^{*}_{i}$ and each line $\v B_{i}$ as two points $\v X_{i}$, $\v X^{*}_{i}$ (Fig. \ref{fig:ramalingam3D}).

Given that both $\v P_{i} = \m A \v X_{i}$  and $\v P^{*}_{i} = \m A \v X^{*}_{i}$ are contained in planes $\v \Pi_{i}$ and $\v \Pi^{*}_{i}$, each line-line correspondence ($\v L_{i}$,$\v B_{i}$) puts 4 linear constraints on $\m A$
\begin{align}
\tr{\v \Pi_{i}}{\m A} \v X_{i} = 0 \label{eq:linear3DUS1}\\
\tr{\v \Pi^{*}_{i}}{\m A} \v X_{i} = 0 \label{eq:linear3DUS2} \\
\tr{\v \Pi_{i}}{\m A} \v X^{*}_{i} = 0 \label{eq:linear3DUS3} \\
\tr{\v \Pi^{*}_{i}}{\m A} \v X^{*}_{i} = 0 \label{eq:linear3DUS4}
\end{align}

Note that the same reasoning can be applied to any combination of plane, point, and line correspondences (planes are defined by 3 points, and points are defined as the intersection of 3 planes), and thus the remainder of this section equally applies to these problems as well. 

\subsection{Linear Solution}
\label{sec:linearsol}

The similarity matrix $\m A$ has 13 linear parameters and for $N$ line-line correspondences we can stack instances of the equations \ref{eq:linear3DUS1} -- \ref{eq:linear3DUS4}) to form a linear system with $4N$ equations and 13 unknowns. This linear system can be solved with SVD decomposition using at least 3 correspondences, determining $\m A$ up to a scale factor. The correct scale of $\m A$ can be recovered by setting the homogeneous parameter to 1. Note, however, that with noisy line measurements equation \ref{eq:similarityNonLin} is not satisfied and thus the linear solution for $\m A$ is generally not a similarity. The linear estimation can be projected to a similarity using QR decomposition of matrix $\m S$ and forcing its upper triangular component to be a scaled identity matrix (using the mean of its diagonal elements as scale $s$)
\begin{equation}
\m S = s\m R = \m R \begin{pmatrix}
s & 0 & 0 \\
0 & s & 0 \\
0 & 0 & s \\
\end{pmatrix}
\end{equation}

\subsection{Minimal Solution}
\label{sec:minimalSol}

Equation \ref{eq:similarityNonLin} puts 5 quadratic constraints on matrix $\m A$ and therefore only 7 linear constraints are required to compute its 13 parameters. This can be achieved with a minimum of 2 line-line correspondences. Note that with 2 correspondences we have 8 linear constraints. To solve the problem minimally we should either discard one of the linear equations or partially solve the complete linear system, leaving 6 up to scale unknowns undetermined. We found the latter option to be numerically more stable. The linear system with 7 equations and 13 unknowns is partially solved using SVD decomposition, generating a 6D solution subspace for $\m A$
\begin{equation}
\label{eq:Asubspace}
\m A = a \m A_{a} + b \m A_{b} + c \m A_{c} + d \m A_{d} + e \m A_{e} + f \m A_{f}
\end{equation}
where $a$, $b$, $c$, $d$, $e$, $f$ are the remaining 6 unknowns.

Equation \ref{eq:similarityNonLin} can be written as the following system of 10 quadratic equations
\begin{equation} 
\label{eq:polySystem}
\begin{aligned}[c]
\tr{\v c_{1}} \v c_{1} - \tr{\v c_{2}} \v c_{2} &= 0 \\
\tr{\v c_{1}} \v c_{1} - \tr{\v c_{3}} \v c_{3} &= 0  \\
\tr{\v c_{1}} \v c_{2} &= 0  \\
\tr{\v c_{1}} \v c_{3} &= 0  \\
\tr{\v c_{2}} \v c_{3} &= 0 
\end{aligned}
\qquad \qquad \qquad
\begin{aligned}[c]
\tr{\v r_{1}} \v r_{1} - \tr{\v r_{2}} \v r_{2} &= 0 \\
\tr{\v r_{1}} \v r_{1} - \tr{\v r_{3}} \v r_{3} &= 0 \\
\tr{\v r_{1}} \v r_{2} &= 0 \\
 \tr{\v r_{1}} \v r_{3} &= 0 \\
 \tr{\v r_{2}} \v r_{3} &= 0 
\end{aligned}
\end{equation}

where $\v c_{i}$ is the $i$th column of $\m S$ and $\v r_{i}$ is the $i$th row of $\m S$. Substituting equation \ref{eq:Asubspace} into equation \ref{eq:polySystem} generates a system of 10 quadratic equations in the 6 unknowns $a$, $b$, $c$, $d$, $e$, $f$. Note that this polynomial system is the same solved in \cite{Ventura2014} for the Generalised Pose and Scale Problem.

This polynomial system is solved with the action matrix method \cite{Byrod2009}. Since the quadratic constraints determine $\m A$ up to scale we set $f = 1$. We expand the polynomial system by multiplying all equations by $a$, $b$, $c$, $d$ and form a cubic system with 47 linearly independent equations and 55 monomials. Using LU decomposition, we reduce the system to 5 equations in 13 monomials 
\begin{equation}
\begin{pmatrix}\m C_{5 \times 5} & \m B_{5 \times 8} \end{pmatrix}
\begin{pmatrix} \v m_{C} \\ \v m_{B}\end{pmatrix}
= 0
\end{equation}
with
\begin{align}
\v m_{C} &= \tr{\begin{pmatrix} b^3 & ab^2 & be & bd & bc \end{pmatrix}} \\
\v m_{B} &= \tr{\begin{pmatrix} b^2 & ab & e & d & c & b & a & 1\end{pmatrix}}
\end{align}
When a polynomial system is presented in this format, it can be solved with the action matrix method if matrix $\m C_{5 \times 5}$ is invertible and also if there is a monomial $w$ such that $w \v m_{B}$ is a linear combination of $\v m_{B}$. In our calibration problem $\m C_{5 \times 5}$ is generally invertible, and for $w = b$ we can build a $8 \times 8$ matrix $\m M$ such that
\begin{equation}
\m M \v m_{B} = b \v m_{B}
\end{equation}
The 8 solutions to $\v m_{B}$ that verify this constraint are the eigen vectors of $\m M$, from which we can extract 8 solutions for $a$, $b$, $c$, $d$, $e$ and recover 8 solutions for $\m A$ using equation \ref{eq:Asubspace}. The correct scale of $\m A$ is recovered in the same way as explained in section \ref{sec:linearsol}.

\section{2D US Calibration Solution}
\label{2DUS}

\begin{figure}[tp]
\centering
\subfigure[]{
\includegraphics[width=0.47\textwidth]{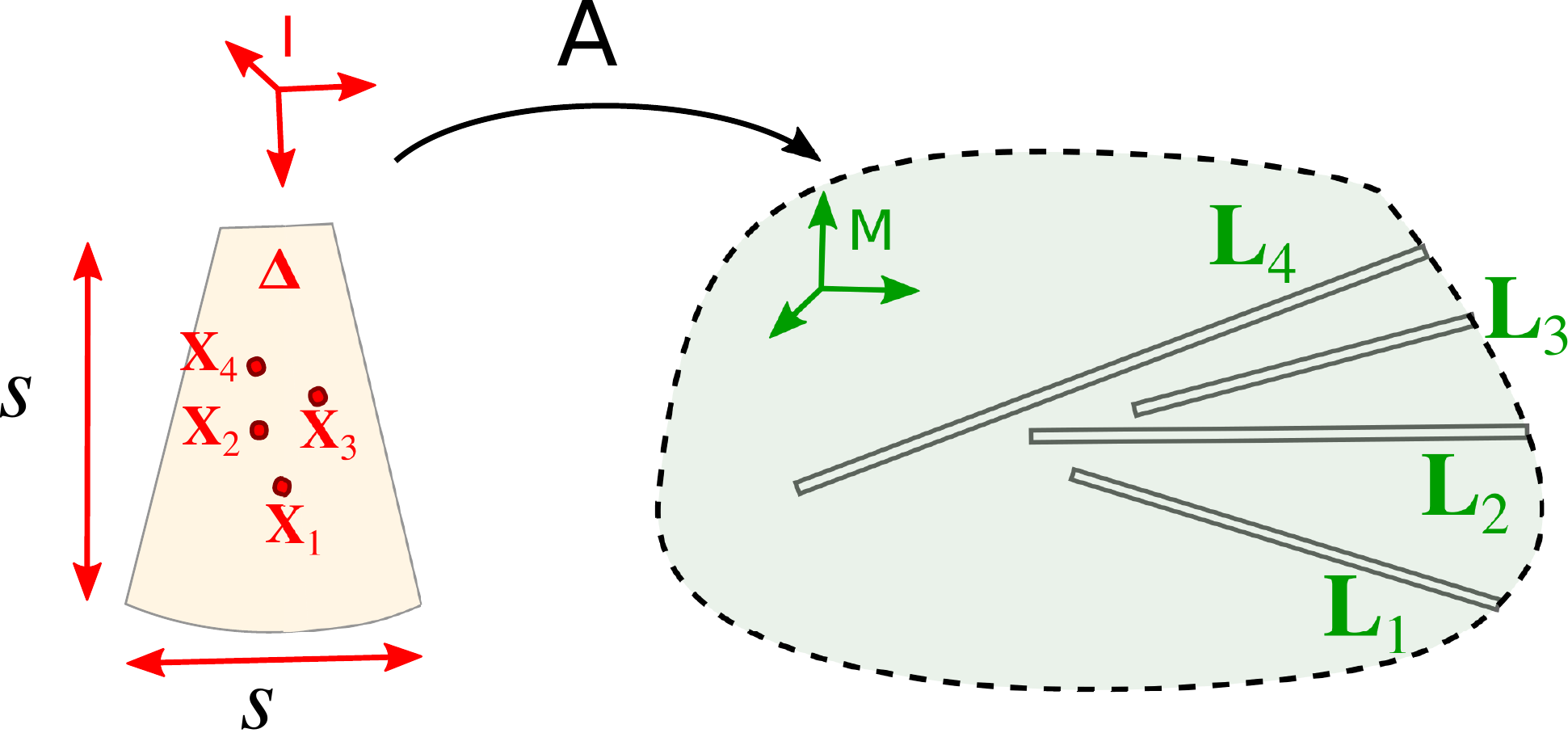}
\label{fig:pointline2D}
}
\subfigure[]{
\includegraphics[width=0.47\textwidth]{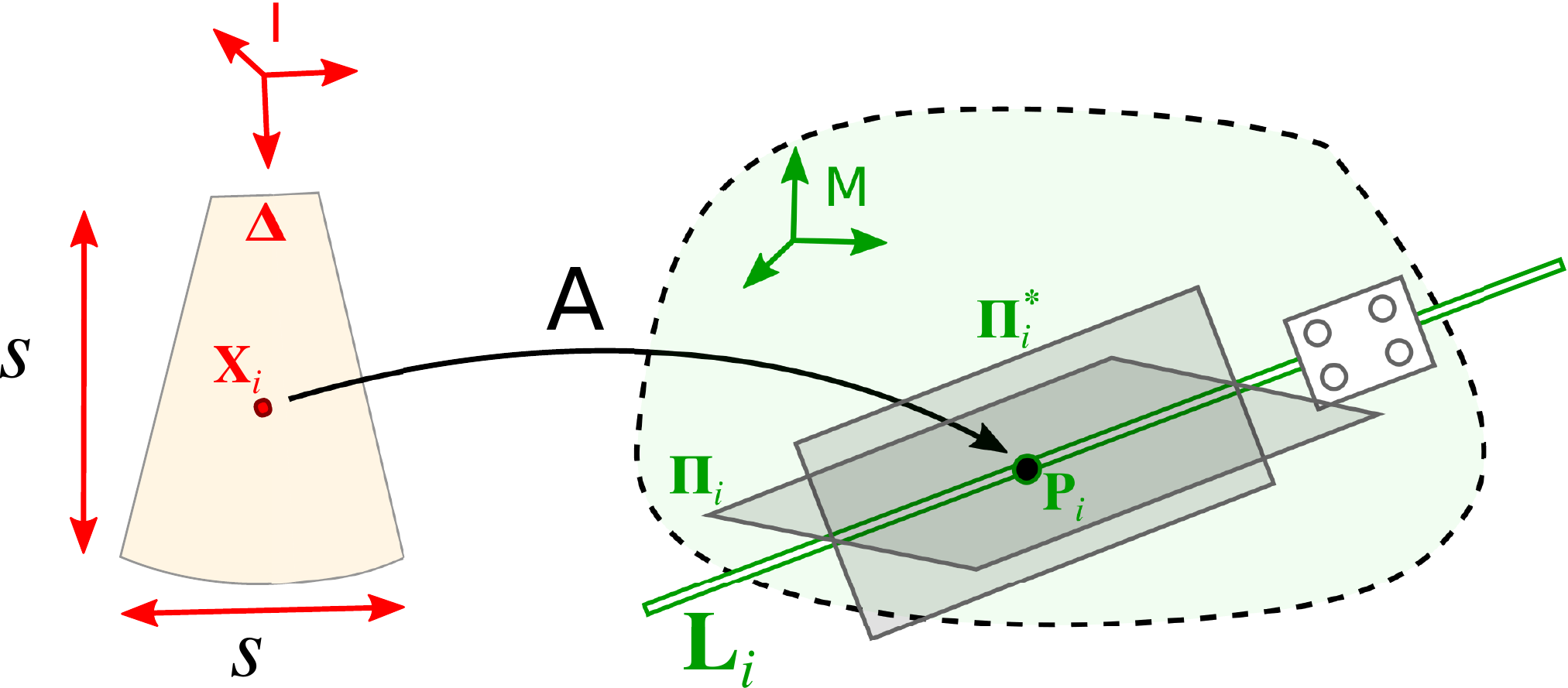}
\label{fig:ramalingam2D}
}
\caption{(a) The 2D US calibration is formulated as the similarity registration between co-planar points $\v P_{i}$ and lines $\v L_{i}$; (b) Each line $\v L_{i}$ can be re-defined as two intersecting planes $\v \Pi_{i}$, $\v \Pi^{*}_{i}$.}
\end{figure}

If we consider the same calibration problem with a curvilinear 2D US probe instead, each needle acquisition is detected as a single point $\v X_{i}$ that belongs to the US scanning plane. For the sake of continuity with the previous section, we still treat the image coordinates $\v X_{i}$ of the 2D US as 3D co-planar points, for an arbitrarily fixed scanning plane $\v \Delta$. Note that calibrating a curvilinear 2D US aims at determining the same 7 parameters as in the 3D US case. Therefore the calibration problem becomes the 3D similarity registration between a set of co-planar points $\v X_{i}$ and a set of lines $\v L_{i}$ (\ref{fig:pointline2D}).

Each point-line correspondence puts 2 linear constraints on matrix $\m A$ (equation \ref{eq:linear3DUS1} and \ref{eq:linear3DUS2}), and therefore this problem can be minimally solved using 4 point-line correspondences. The same minimal solution described in section \ref{sec:minimalSol} can be used in this case, since the co-planarity of points $\v X_{i}$ is not a degenerate configuration. We observed that some particular choices for the scanning plane $\v \Delta$ (e. g. $z=0$) result in matrix $\m C_{5 \times 5}$ being singular and thus the polynomial system becomes numerically unstable. We found out through simulation that defining $\v \Delta$ as the plane $z=k$, with $k>0$ generally produces an invertible matrix $\m C_{5 \times 5}$ and the polynomial system is solvable.

On the other hand, the linear solution described in section \ref{sec:linearsol} will not solve the 2D US problem, as the system will always be rank deficient. This can only be achieved with the additional elimination of parameters in the linear equations. These simplifications also lead to an alternative minimal solution for the 2D US case. Both methods are described in the remainder of this section.

\subsection{Linear Solution}
\label{sec:linearSol2D}

If we define the scanning plane $\v \Delta$ as $z=0$, the 2D US points have the format $\v X_{i} = \tr{\begin{pmatrix}
x_{i} & y_{i} & 0 & 1
\end{pmatrix}}$ and the linear equations do not put any constraints on the third column of $\m S$. The linear equations for each acquisition become
\begin{align}
\tr{\v \Pi_{i}} \bar{\m A} \tr{\begin{pmatrix}
x_{i} & y_{i} & 1
\end{pmatrix}} &= 0 \\
\tr{\v \Pi^{*}_{i}} \bar{\m A} \tr{\begin{pmatrix}
x_{i} & y_{i} & 1
\end{pmatrix}} &= 0 
\end{align}
with
\begin{align}
\bar{\m A} &= \begin{pmatrix}
\bar{\m S} & \v t \\
0 & 1
\end{pmatrix} \\
\bar{\m S} &= \begin{pmatrix}
\v c_{1} & \v c_{2}
\end{pmatrix}
\end{align}
where $\v c_{1}$ and $\v c_{2}$ are the first two columns of $\m S$. The linear system is thus reduced to 10 unknown parameters and can be solved with a minimum of 5 point-line correspondences. Note that analogously to the 3D US case (equation \ref{eq:similarityNonLin}), $\bar{\m A}$ must verify the following constraint
\begin{equation}
\label{eq:similarityNonLin2D}
\bar{\m S} \tr{\bar{\m S}} = \begin{pmatrix}
s^{2} & 0 \\
0 & s^{2}
\end{pmatrix}
\end{equation}
and with noisy measurements the linear solution must be forced to this format using its QR decomposition
\begin{equation}
\bar{\m S} = \m R \begin{pmatrix}
s & 0 \\
0 & s \\
0 & 0
\end{pmatrix}
\end{equation}
The third column of $\m S$ can then be extracted by multiplying the third column of rotation $\m R$ by $s$.

\subsection{Alternative Minimal Solution (2D US only)}
\label{sec:minimalSol2D}

This problem can be minimally solved with 7 linear constraints (4 point-line correspondences). Since in this case there are only 10 linear parameters, we can generate a 3D linear solution subspace
\begin{equation}
\label{eq:Asubspace2D}
\bar{\m A} = a \bar{\m A}_{a} + b \bar{\m A}_{b} + c \bar{\m A}_{c}
\end{equation}

Equation \ref{eq:similarityNonLin2D} is re-written as the following system
\begin{align} 
\label{eq:polySystem2D}
\begin{split}
\tr{\v c_{1}} \v c_{1} - \tr{\v c_{2}} \v c_{2} &= 0 \\
\tr{\v c_{1}} \v c_{2} &= 0
\end{split}
\end{align}
Substituting equation \ref{eq:Asubspace2D} into equation \ref{eq:polySystem2D} we generate a system of 2 quadratic homogeneous equations in the 3 unknowns $a$, $b$, $c$. Using the same procedure from section \ref{sec:minimalSol} we use monomial multiplication and LU decomposition to re-write this system as
\begin{equation}
\begin{pmatrix}
\m C_{2 \times 2} & \m B_{2 \times 4}
\end{pmatrix} \begin{pmatrix}
\v m_{C} \\ \v m_{B}
\end{pmatrix} = 0
\end{equation}
with
\begin{align}
\v m_{C}  &= \tr{\begin{pmatrix}
ab^{2} & b^{2}
\end{pmatrix}} \\
\v m_{B}  &= \tr{\begin{pmatrix}
 ab & b & a & 1
\end{pmatrix}}
\end{align}

We solve this system using eigen decomposition of the action matrix, yelding up to 4 solutions.

\section{Degenerate Cases}

The degenerate configurations for both 3D US and 2D US calibration are closely related to the ones described for the pose and scale problem \cite{Ventura2014}. If the needle is moved without rotation (lines $\v L_{i}$ are parallel) there is an ambiguity in translation. This implies that fixing the needle and scanning with the US probe in different positions is a degenerate case, however, the inverse scenario of fixing the US probe while moving the needle is generally not a degenerate case. If the needle motion is a pure rotation around itself (lines $\v L_{i}$ intersect in a single point) there is an ambiguity in scale. This is analogous to pose estimation with monocular pinhole cameras. If lines $\v L_{i}$ are co-planar, the point detections $\v X_{i}$ of a 2D US are co-linear and there is a rotation ambiguity around the axis defined by these points. This, however, is not generally a degenerate case in 3D US calibration unless only 2 line correspondences are available, since it falls under either one of the two previously mentioned cases. Therefore, the similarity between two sets of co-planar lines can only be estimated from a minimum of 3 correspondences.

\section{Iterative refinement}


The closed-form solutions can be refined with Levenberg-Marquardt iterative optimization \cite{Marquardt1963}, however, there is no consensus on the most appropriate residue metric for 3D line registration \cite{Bartoli2001}. In all our experiments we perform iterative refinement by minimizing the euclidean orthogonal distance between the 3D lines $\v L_{i}$ and the projected 3D points from the US image $\v P_{i} = \m A \v X_{i}$. The refined solution is parametrised by the translation $\v t$, 3 rotation parameters ($\v R$ is represented as a unit norm quaternion), and the scale factor $s$. For the 3D US the optimization problem is
\begin{equation}
	\min_{\m R, \v t, s} \sum_{i = 1}^{N} d(\v L_{i},\v P_{i})^{2} + d(\v L_{i},\v P^{*}_{i})^{2} 
\end{equation}
where $d(\v L_{i},\v P_{i})$ represents the euclidean distance between line $\v L_{i}$ and point $\v P_{i}$. For the 2D US problem the last term of the minimization is ignored.

\section{Experimental Results}

We test the calibration algorithms both in simulation and with real data. For the 3D US calibration we test the linear solution from section \ref{sec:linearsol} (\textbf{3line3D}) and the minimum solution from section \ref{sec:minimalSol} (\textbf{2line3D}), while for the 2D US we test the linear solution from section \ref{sec:linearSol2D} (\textbf{5point2D}), the general minimal solution from section \ref{sec:minimalSol} (\textbf{2line3D}), and the simplified minimum solution from section \ref{sec:minimalSol2D} (\textbf{4point2D}). All algorithms are tested within a RANSAC framework \cite{Fischler1981} with outlier threshold of 5mm, followed by iterative refinement.

\begin{figure}[t]
\centering
\subfigure[]{\includegraphics[width=0.4\textwidth]{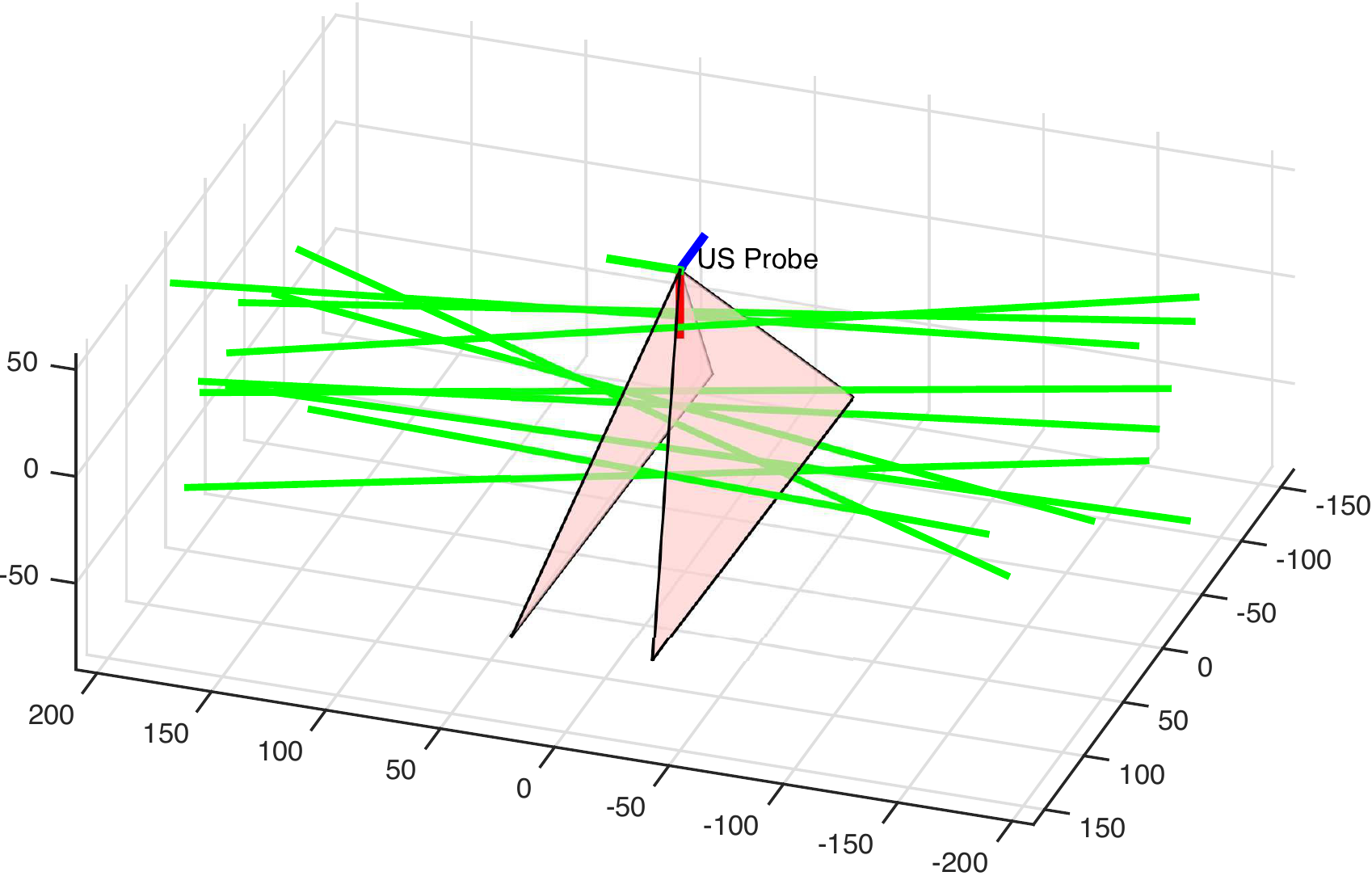}
}
\hspace{0.6cm}
\subfigure[]{\includegraphics[width=0.17\textwidth]{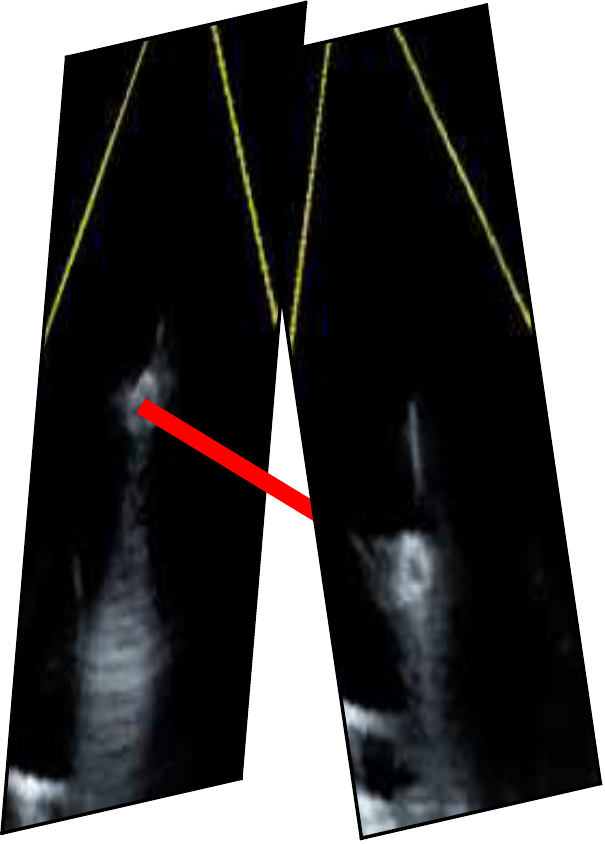}
}
\hspace{0.6cm}
\subfigure[]{\includegraphics[width=0.17\textwidth]{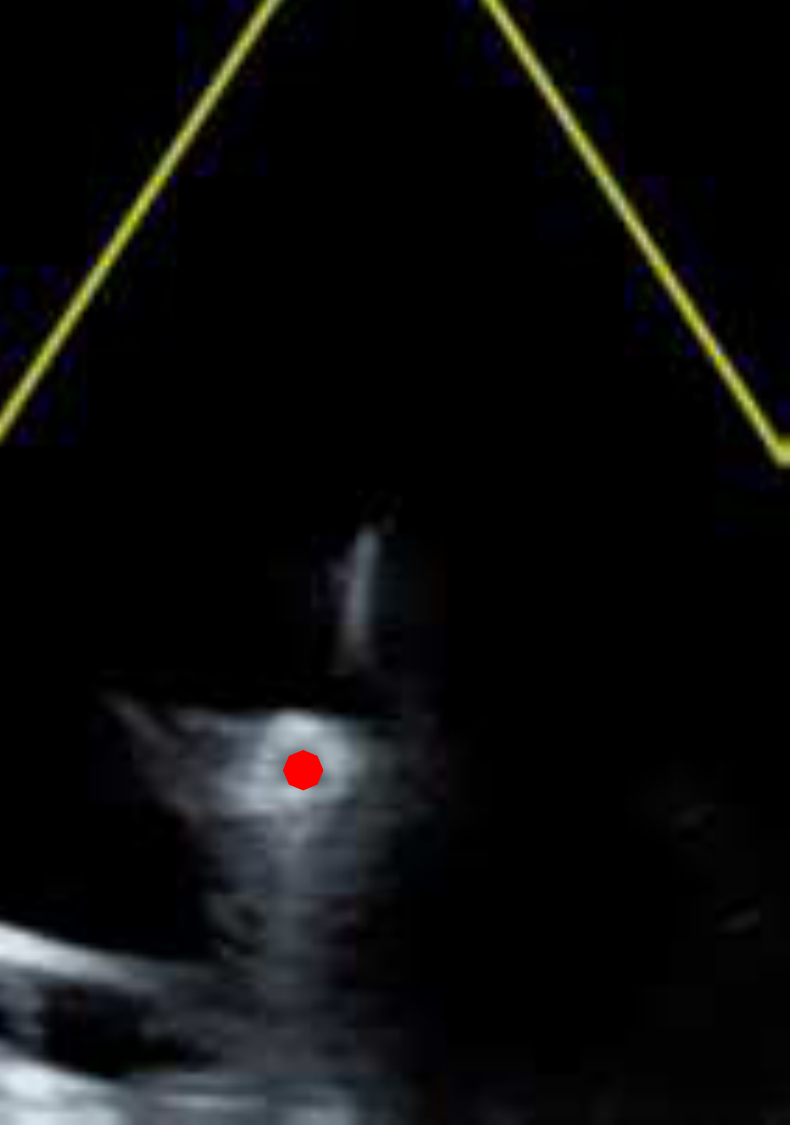}
}
\caption{The needle detections in 3D are obtained by sampling 2D slices: (a) Simulated set-up with a fixed 3D US scanner and random needle poses in green; (b) 3D US line acquisition; 2D US point acquisition.}
\label{fig:usDetections}
\end{figure}

With both synthetic and real data, the 3D US lines $\v B_{i}$ are obtained by sampling points from 2D slices with different angles (Fig. \ref{fig:usDetections}). This is is a practical solution since we can directly define the points $\v X_{i}$ and $\v X^{*}_{i}$ to input in equations \ref{eq:linear3DUS1} to \ref{eq:linear3DUS4}. The needle tracking measurements (lines $\v L_{i}$) are converted to two intersecting planes $\v \Pi_{i}$ and $\v \Pi^{*}_{i}$ such that $\v \Pi_{i}$ intersects both $\v L_{i}$ and the origin of the US attached marker reference frame $\m M$, and $\v \Pi_{i}^{*}$ is orthogonal to $\v \Pi{i}$. Note that this approach degenerates when the needle is aligned with the origin of $\m M$ and therefore this should be taken into account when positioning the needle during calibration. 

All plots in this section are represented with Matlab boxplot function: the central mark is the median, the box limits are the 25th and 75th percentiles, the whiskers are the maximum and minimum inliers, and individual crosses are outliers. 

\subsection{Simulation}

\begin{figure}[tp]
\centering
\subfigure[]{\includegraphics[width=0.31\textwidth]{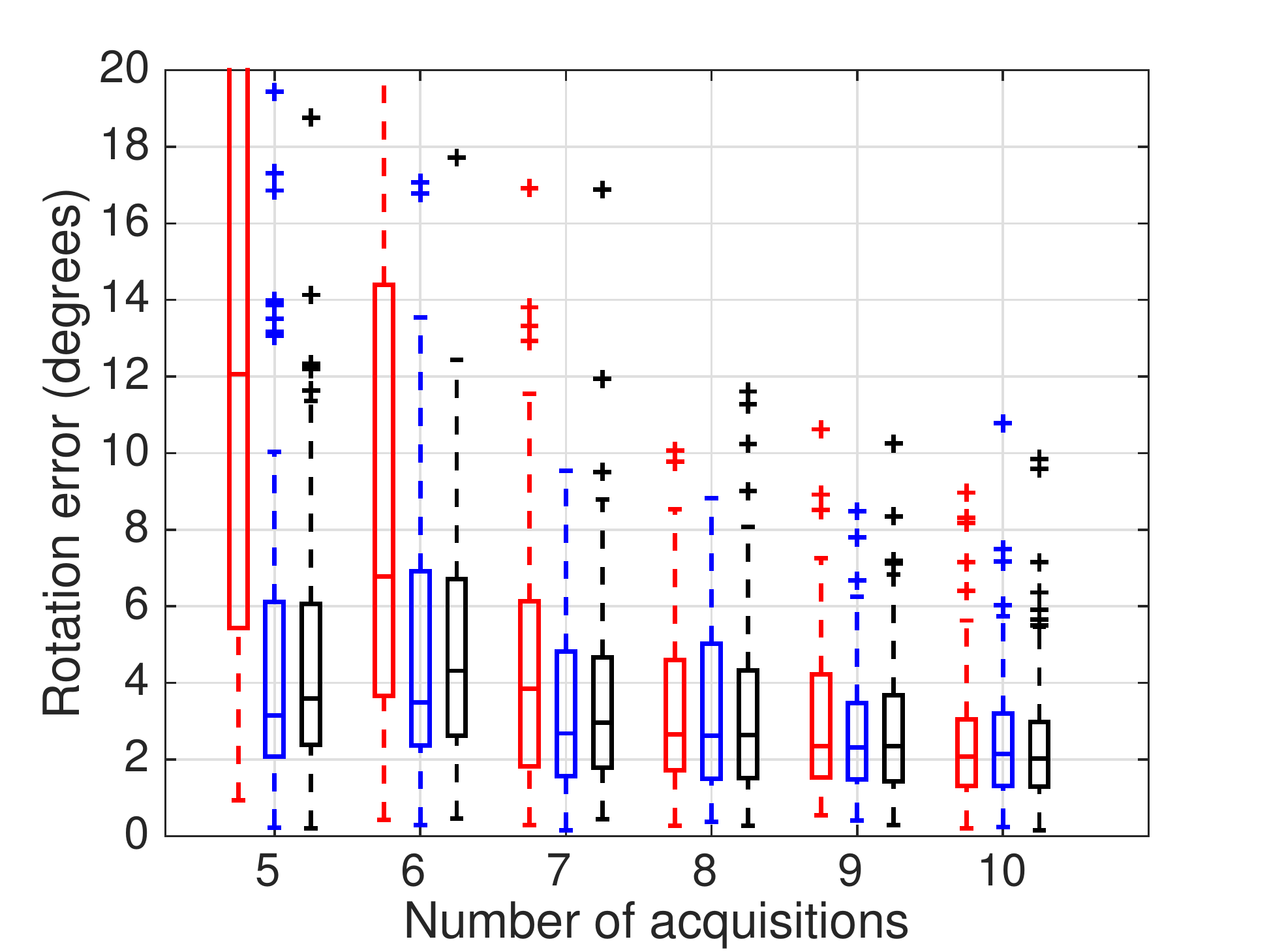}
}
\subfigure[]{\includegraphics[width=0.31\textwidth]{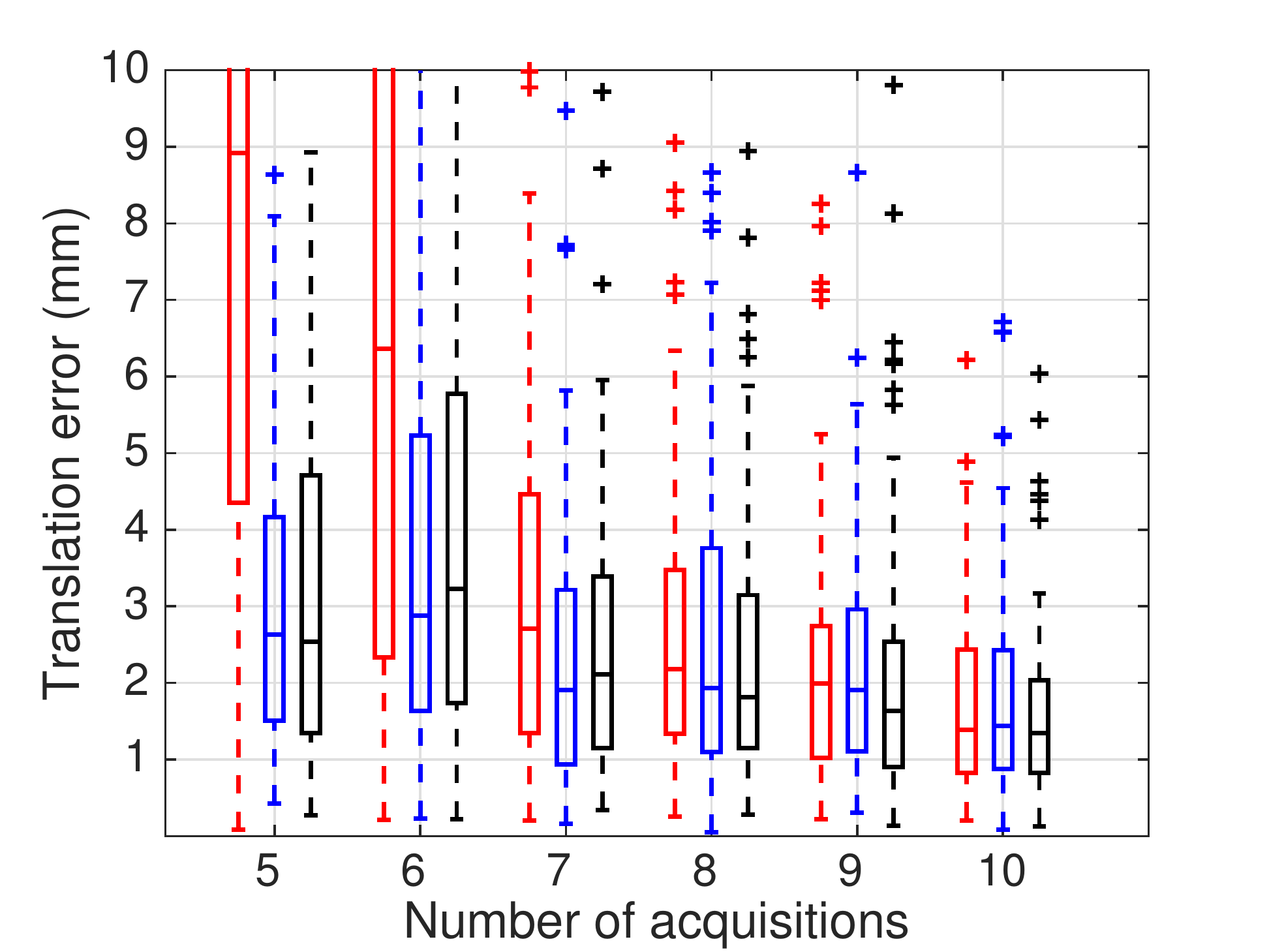}
}
\subfigure[]{\includegraphics[width=0.31\textwidth]{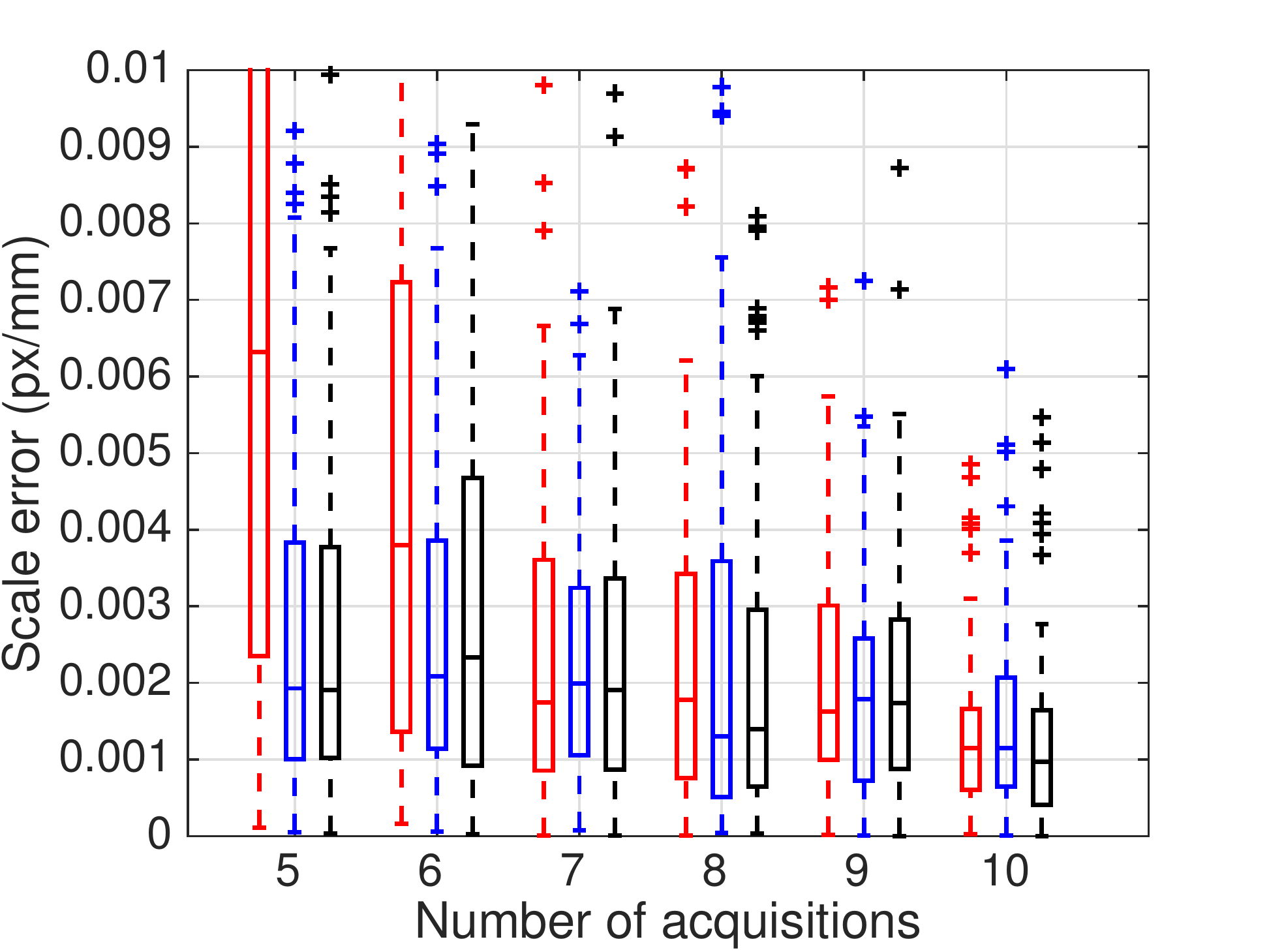}
}
\subfigure{\includegraphics[width=0.5\textwidth]{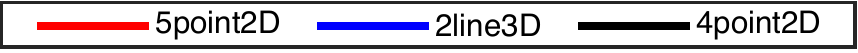}
}
\caption{2D US error distributions with synthetic data}
\label{fig:2DSimResults}
\end{figure}

\begin{figure}[t]
\centering
\subfigure[]{\includegraphics[width=0.31\textwidth]{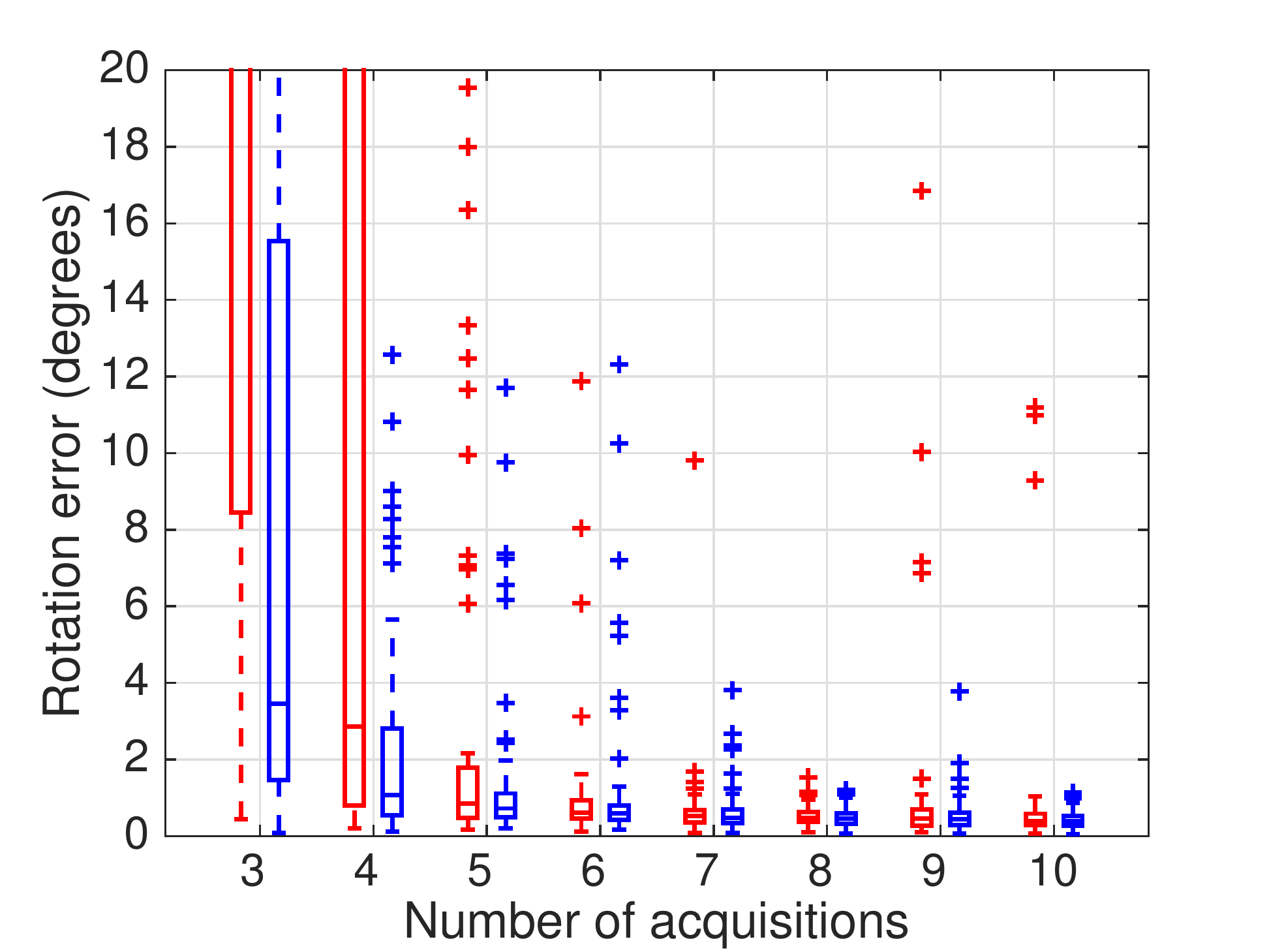}
}
\subfigure[]{\includegraphics[width=0.31\textwidth]{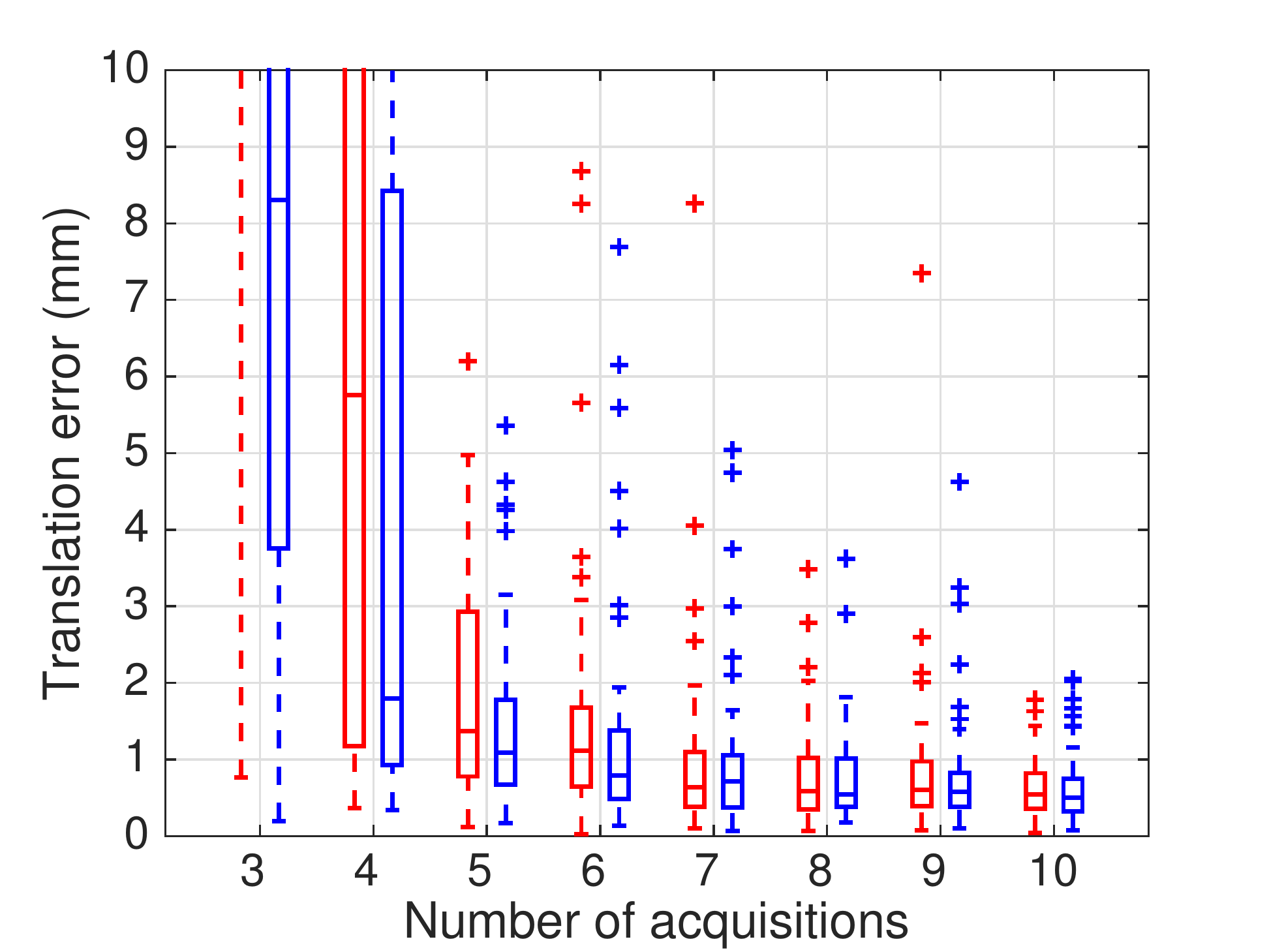}
}
\subfigure[]{\includegraphics[width=0.31\textwidth]{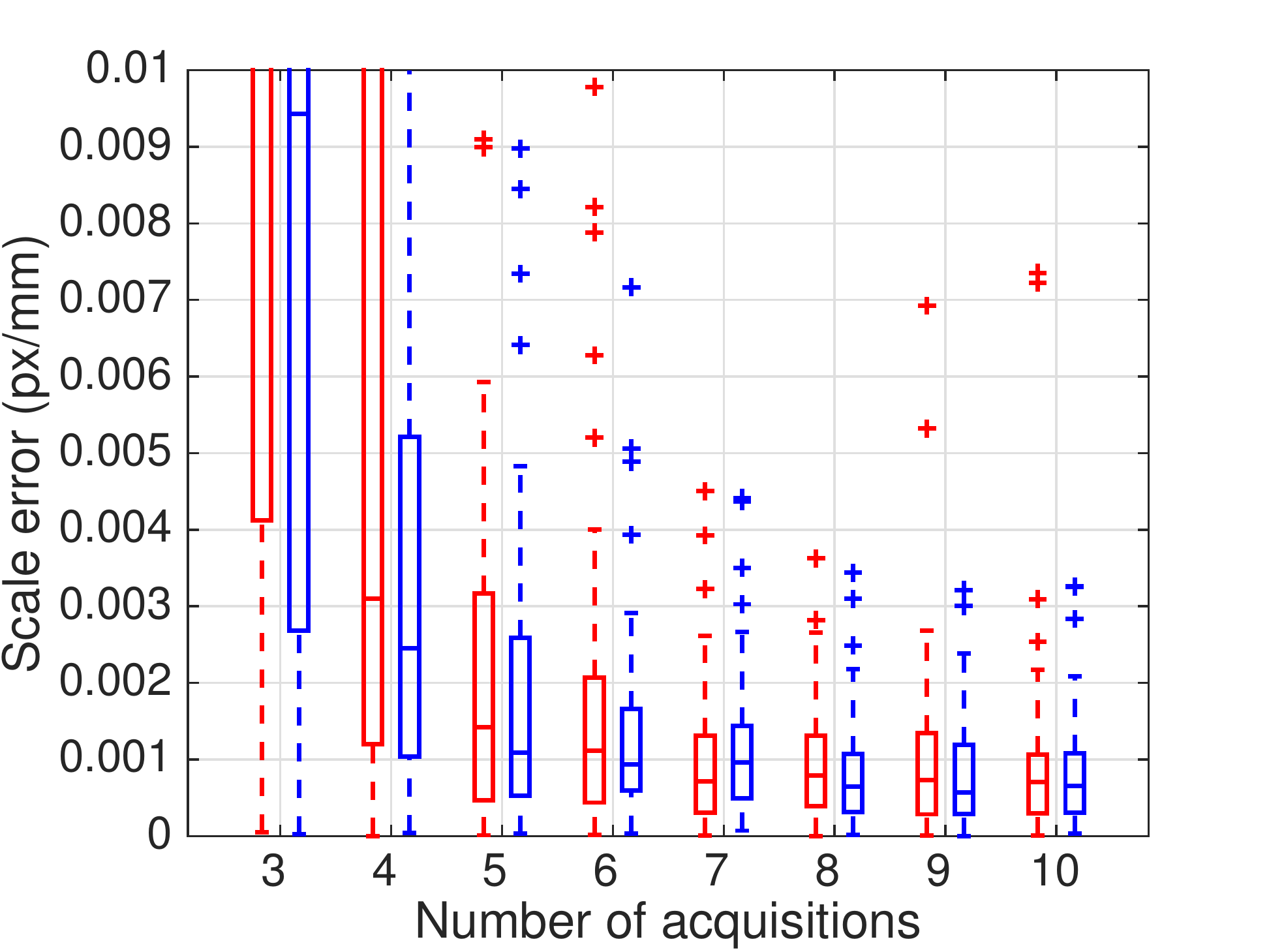}
}
\subfigure{\includegraphics[width=0.35\textwidth]{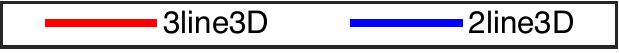}
}
\caption{3D US error distributions with synthetic data}
\label{fig:3DSimResults}
\end{figure}

We simulate a 2D/3D US probe with a scale factor $s=0.24$ in a fixed position. 50 line segments with 400mm are generated at random poses within the field of view of the US. Gaussian noise is added to the US points along the 2D slices ($\sigma = 1$ pixel) and also to the extreme points of the line segments $\v L_{i}$ ($\sigma = 1$mm) to simulate tracking error. In each trial, we calibrate the US by sampling $N$ random line segments. $N$ varies between 3 and 10 for the 3D US case, and between 5 and 10 for the  2D US case. For each value $N$ we perform 100 trials. The calibration results are compared against of rotation ($\m R_{GT}$), translation ($\v t_{GT}$), and scale factor ($s_{GT}$) and are presented in figures \ref{fig:2DSimResults} and \ref{fig:3DSimResults}. The rotation error is measured as the angle displacement of the residual rotation $\tr{\m R} \m R_{GT}$, the translation error as $||\v t_{GT} - \v t||$ and the scale error as $|s_{GT} - s|$. As expected, the minimal solutions perform better than the linear solutions with a low number of input acquisitions and converge to the same result as the number of acquisitions grow. Also not surprisingly, 3D US performs better that 2D US for the same number of acquisitions, given that it has twice as much linear constraints. In the case of 2D US, the two alternative minimum solutions have similar performance. 

\subsection{Real Data}

\begin{figure}[tp]
\centering
\subfigure[]{\includegraphics[width=0.48\textwidth]{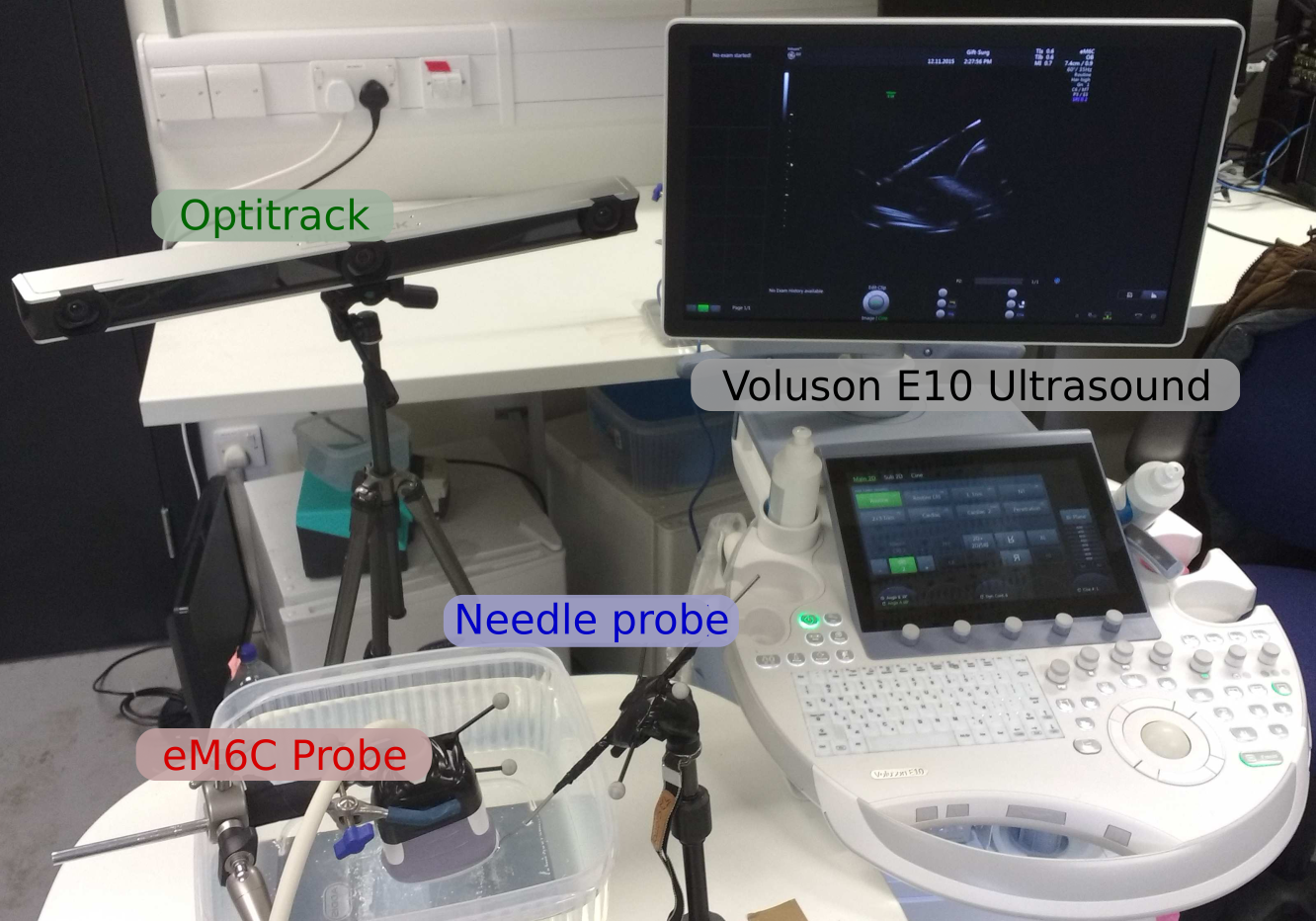}
\label{fig:2D3DCalibSetup}
}
\hspace{0.1\textwidth}
\begin{minipage}[b]{0.33\textwidth}
\subfigure{
\includegraphics[width=1.0\textwidth]{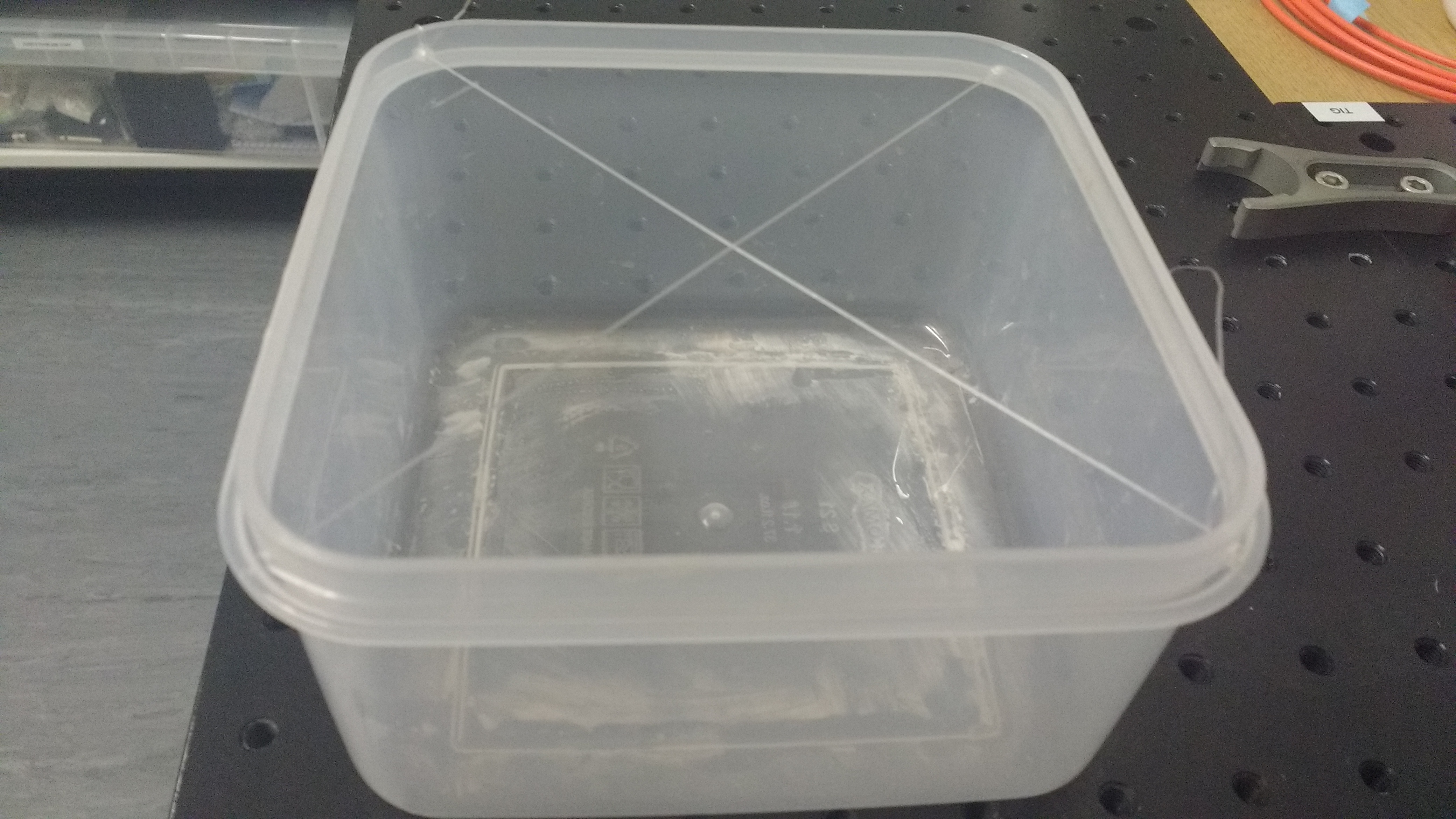} 
}
\addtocounter{subfigure}{-1}
\subfigure[]{
\includegraphics[width=1.0\textwidth]{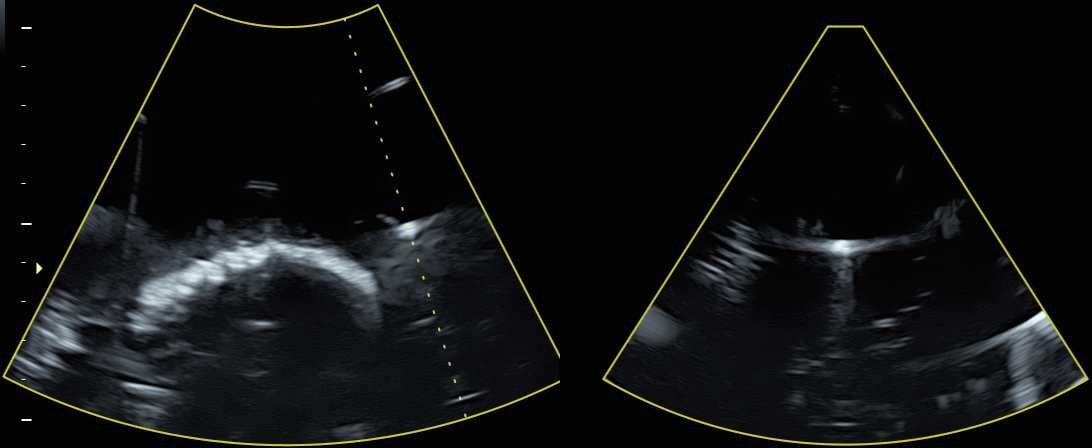}
\label{fig:ValidSetup}
}
\end{minipage}
\caption{(a) Calibration set-up with a tracked 3D US probe and a tracked needle; (b) Validation using a cross wire pattern that defines a known 3D point in the tracker reference frame.}
\end{figure}

Our calibration method is tested using the set-up displayed in Fig. \ref{fig:2D3DCalibSetup} that includes a GE Voluson E10 machine with a eM6C probe (3D US) and a 333 mm length metal needle. Both instruments are tracked by the infrared camera system Optitrack V120 Trio. Experiments were conducted in a container filled with water at room temperature. We use the same probe for both for 2D US and 3D US data acquisition. For the 2D US we just choose a single 2D slice from the 3D US volume at a specified angle. The needle is manually segmented as a point in each 2D slice. Unlike in the simulation experiment, in this calibration procedure both the needle and the 3D US probe are moved between acquisitions. 

\begin{figure}[t]
\centering
\begin{minipage}[b]{0.35\textwidth}
\centering
\subfigure{\includegraphics[width=1.0\textwidth]{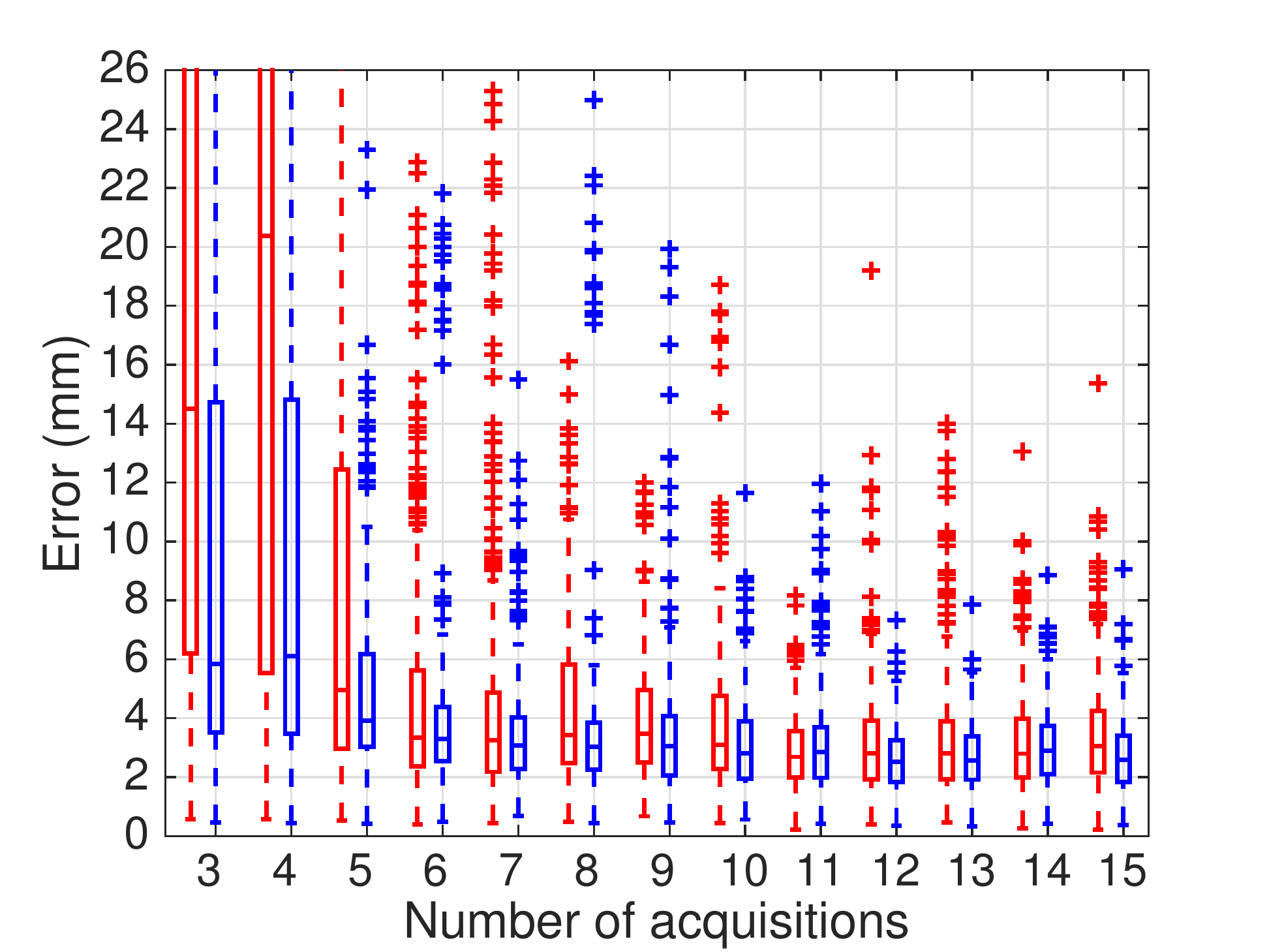}
\label{fig:2DPRA}
}
\addtocounter{subfigure}{-1}
\subfigure[]{\includegraphics[width=0.8\textwidth]{./legend3D-eps-converted-to}
}
\end{minipage}
\begin{minipage}[b]{0.35\textwidth}
\centering
\subfigure{\includegraphics[width=1.0\textwidth]{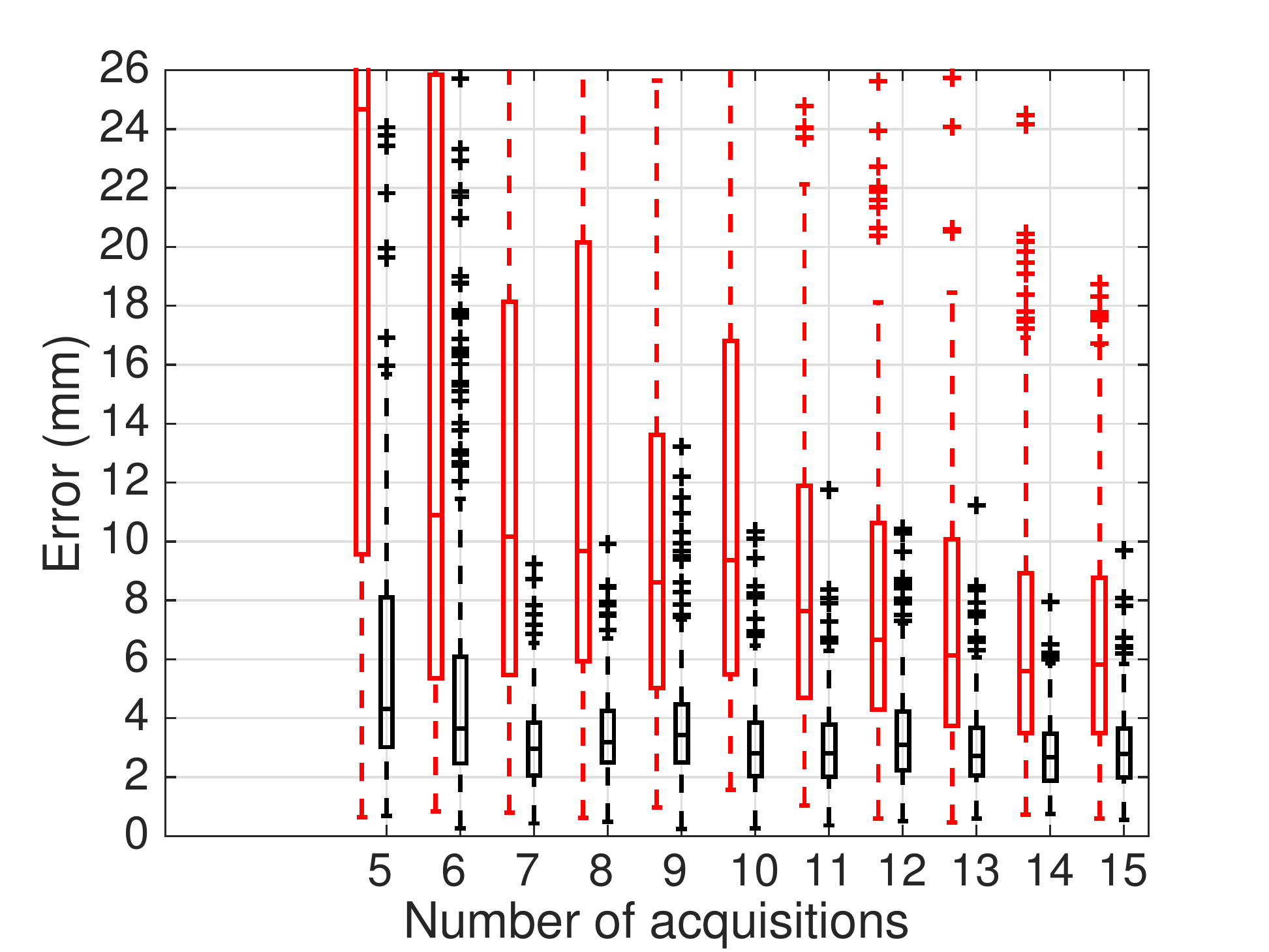}
\label{fig:3DPRA}
}
\addtocounter{subfigure}{-1}
\subfigure[]{\includegraphics[width=0.8\textwidth]{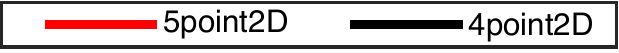}
}
\end{minipage}
\subfigure[]{\includegraphics[width=0.45\textwidth]{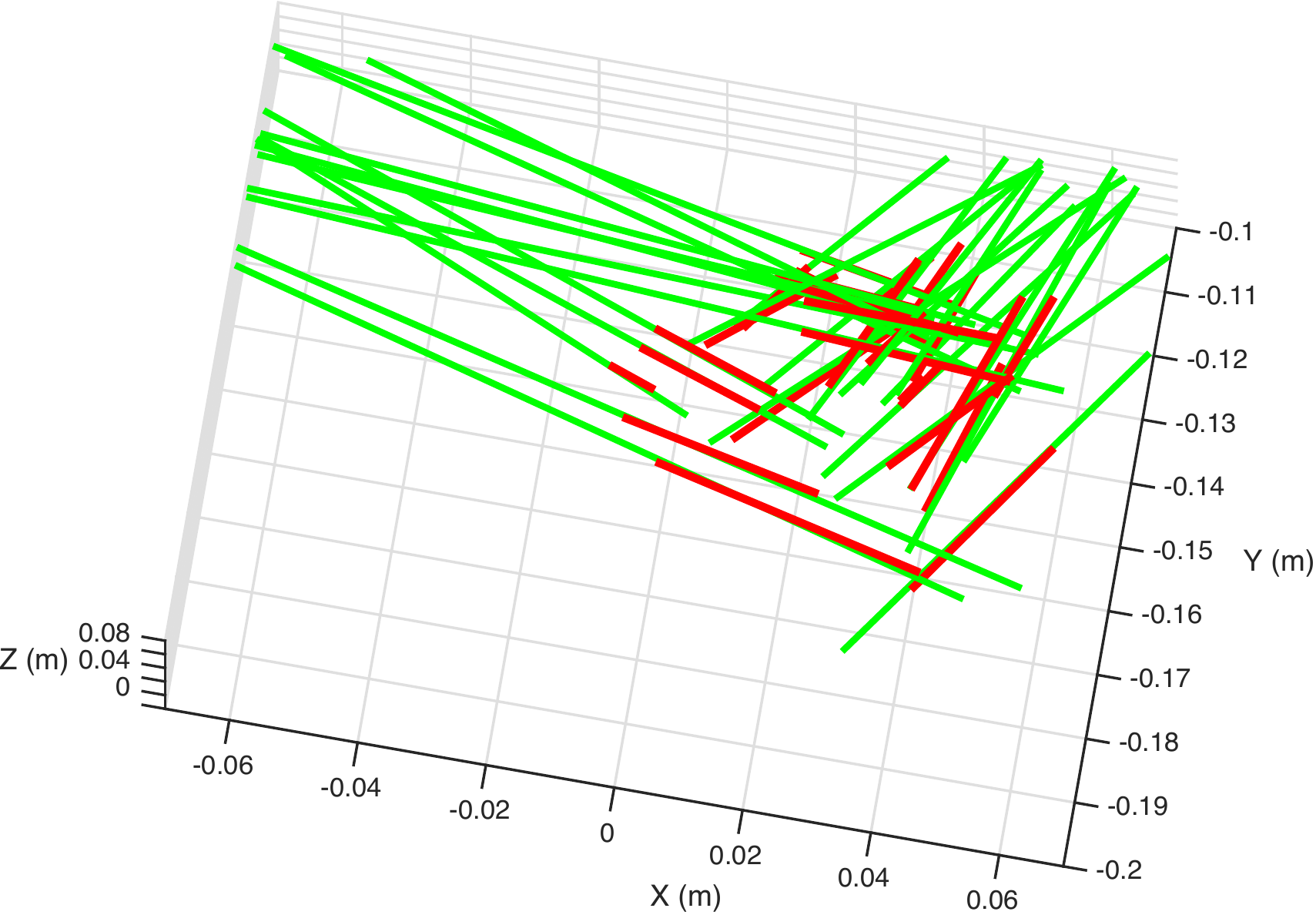}

}
\subfigure[]{\includegraphics[width=0.42\textwidth]{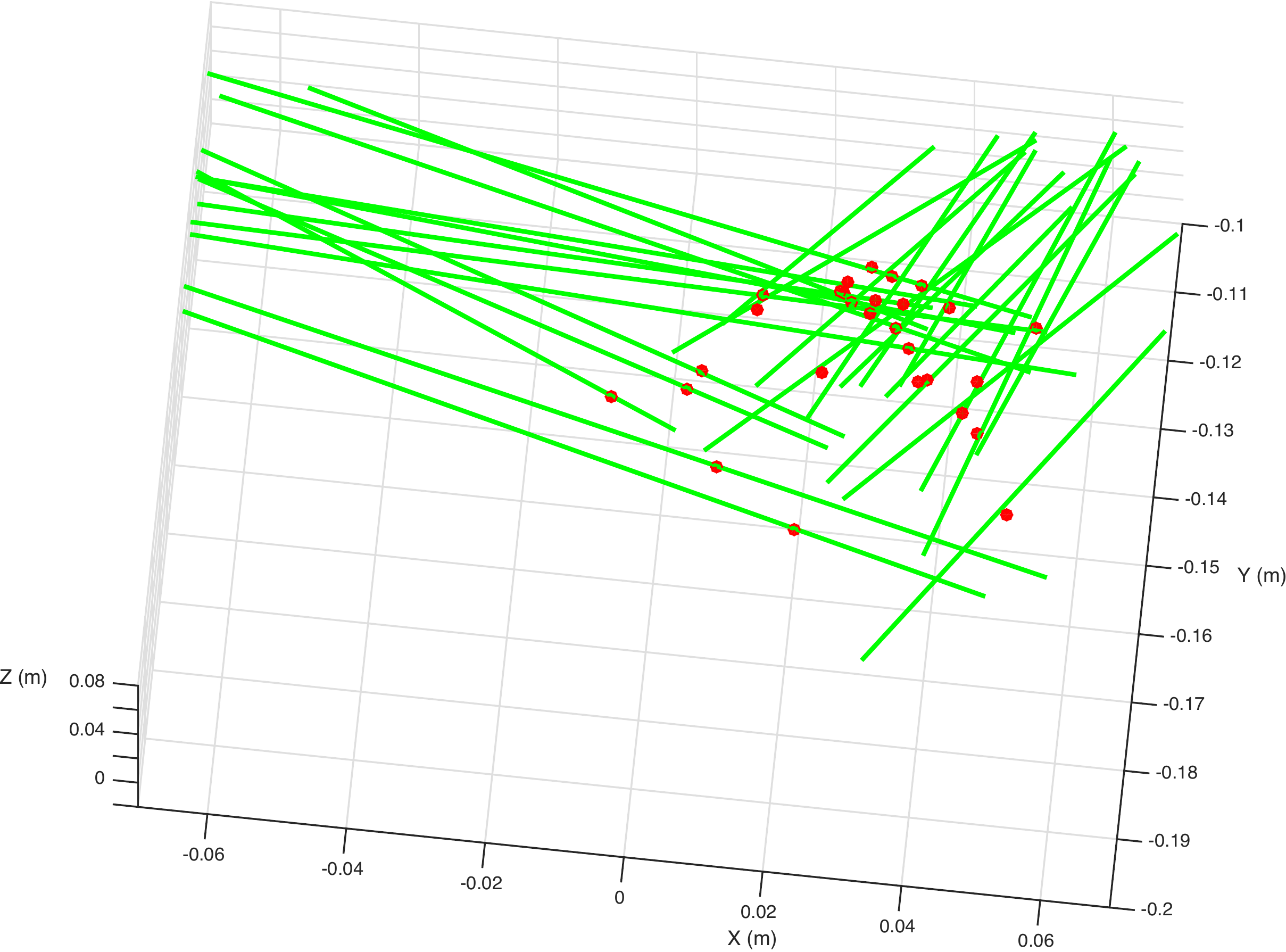}

}
\caption{Validation results: (a) Projection reconstruction accuracy (PRA) with 3D US; (b) PRA with 2D US (c) Sample registration result with 3D US projected lines in red and needle tracking measurements in green (d) Sample registration result with 2D US projected points in red and needle tracking measurements in green}
\end{figure}

To validate the calibration accuracy we use an x-shaped wire phantom \ref{fig:ValidSetup} whose point intersection can be measured as a single point in the US scan. We use this phantom to measure the projection reconstruction accuracy (PRA) of our calibration results,, i. e., the difference in mm between the intersection point $\m A \v X$ according to the calibrated US measurement and the same point $\v P$ measured by the tip of the tracked needle. We performed 10 acquisitions of the wire phantom in order to cover different regions of the US scan. Figs. \ref{fig:3DPRA} and \ref{fig:2DPRA} display the distribution of PRA results for all trials. Each distribution contains 200 error measurements (20 trials $\times$ 10 phantom scans). In the 2D US results we only display the results for one of the minimal solutions, since as we've seen in simulation the results from both approaches are very similar.

The US calibration converges to a solution with with an error between 2 and 3 mm within a total scanning radius of 120 mm. Since in each trial we select random needle poses, close to degenerate configurations can be chosen and result in outlier results. This can be avoided in practice by scanning the needle in a spread region of the US volume, and by exploring all 6 degrees of freedom while moving the needle. Overall, the difference in accuracy between linear and minimal solutions is even more pronounced than in simulation.

\section{Conclusions}

We propose a minimum solution to the similarity registration between two sets of 3D lines, and between co-planar points and 3D lines. These solutions are tested to calibrate a US probe using a tracked line target with both 3D and 2D data. This is useful in medical imaging to guide a biopsy needle during US based interventions. The method can be easily be extended to additional US calibration problems using other types of phantoms, e. g. scanning single plane target leads to the similarity registration between co-planar lines and 3D planes (2D US) or between two sets of 3D planes (3D US). In other computer vision domains this algorithm can potentially be used as an extension of the pose and scale problem to the alignment of line-based and/or plane-based SfM sequences.


\bibliographystyle{splncs}
\bibliography{eccv2016submission}

\begin{thebibliography}{10}

\bibitem{Khamene2005}
Khamene, A., Sauer, F.:
\newblock A novel phantom-less spatial and temporal ultrasound calibration
  method.
\newblock In Duncan, J., Gerig, G., eds.: Medical Image Computing and
  Computer-Assisted Intervention – MICCAI 2005. Volume 3750 of Lecture Notes
  in Computer Science.
\newblock Springer Berlin Heidelberg (2005)  65--72

\bibitem{Brattain2011}
Brattain, L., Howe, R.:
\newblock Real-time 4d ultrasound mosaicing and visualization.
\newblock In Fichtinger, G., Martel, A., Peters, T., eds.: Medical Image
  Computing and Computer-Assisted Intervention – MICCAI 2011. Volume 6891 of
  Lecture Notes in Computer Science.
\newblock Springer Berlin Heidelberg (2011)  105--112

\bibitem{Stoll2012}
Stoll, J., Ren, H., Dupont, P.:
\newblock Passive markers for tracking surgical instruments in real-time 3-d
  ultrasound imaging.
\newblock Medical Imaging, IEEE Transactions on \textbf{31}(3) (March 2012)
  563--575

\bibitem{Mercier2005}
Mercier, L., Langø, T., Lindseth, F., Collins, L.D.:
\newblock A review of calibration techniques for freehand 3-d ultrasound
  systems.
\newblock Ultrasound in Medicine \& Biology \textbf{31}(2) (2005)  143 -- 165

\bibitem{Chen2009}
Chen, T.K., Thurston, A.D., Ellis, R.E., Abolmaesumi, P.:
\newblock A real-time freehand ultrasound calibration system with automatic
  accuracy feedback and control.
\newblock Ultrasound in Medicine \& Biology \textbf{35}(1) (2009)  79 -- 93

\bibitem{Prager1998}
Prager, R., Rohling, R., Gee, A., Berman, L.:
\newblock Rapid calibration for 3-d freehand ultrasound.
\newblock Ultrasound in Medicine \& Biology \textbf{24}(6) (1998)  855 -- 869

\bibitem{Najafi2015}
Najafi, M., Afsham, N., Abolmaesumi, P., Rohling, R.:
\newblock A closed-form differential formulation for ultrasound spatial
  calibration: Single wall phantom.
\newblock Ultrasound in Medicine \& Biology \textbf{41}(4) (2015)  1079 -- 1094

\bibitem{Muratore2001}
Muratore, D.M., Jr, R.L.G.:
\newblock Beam calibration without a phantom for creating a 3-d freehand
  ultrasound system.
\newblock Ultrasound in Medicine \& Biology \textbf{27}(11) (2001)  1557 --
  1566

\bibitem{Hsu2008}
Hsu, P.W., Treece, G.M., Prager, R.W., Houghton, N.E., Gee, A.H.:
\newblock Comparison of freehand 3-d ultrasound calibration techniques using a
  stylus.
\newblock Ultrasound in Medicine \& Biology \textbf{34}(10) (2008)  1610 --
  1621

\bibitem{Najafi2014}
Najafi, M., Afsham, N., Abolmaesumi, P., Rohling, R.:
\newblock A closed-form differential formulation for ultrasound spatial
  calibration: Multi-wedge phantom.
\newblock Ultrasound in Medicine \& Biology \textbf{40}(9) (2014)  2231 -- 2243

\bibitem{Bergmeir2009}
Bergmeir, C., Seitel, M., Frank, C., Simone, R., Meinzer, H.P., Wolf, I.:
\newblock Comparing calibration approaches for 3d ultrasound probes.
\newblock International Journal of Computer Assisted Radiology and Surgery
  \textbf{4}(2) (2009)  203--213

\bibitem{Hummel2013}
Hummel, J., Kaar, M., Hoffmann, R., Bhatia, A., Birkfellner, W., Figl, M.:
\newblock Evaluation of three 3d us calibration methods.
\newblock Proc. SPIE \textbf{8671} (2013)  86712I--86712I--8

\bibitem{Vasconcelos2016}
Vasconcelos, F., Peebles, D., Ourselin, S., Stoyanov, D.:
\newblock Spatial calibration of a 2d/3d ultrasound using a tracked needle.
\newblock International Journal of Computer Assisted Radiology and Surgery
  \textbf{11}(6) (2016)  1091--1099

\bibitem{Zhang2004}
Zhang, Q., Pless, R.:
\newblock Extrinsic calibration of a camera and laser range finder (improves
  camera calibration).
\newblock In: Intelligent Robots and Systems, 2004.(IROS 2004). Proceedings.
  2004 IEEE/RSJ International Conference on. Volume~3., IEEE (2004)  2301--2306

\bibitem{Horaud1995}
Horaud, R., Dornaika, F.:
\newblock Hand-eye calibration.
\newblock The international journal of robotics research \textbf{14}(3) (1995)
  195--210

\bibitem{Thompson2016}
Thompson, S., Stoyanov, D., Schneider, C., Gurusamy, K., Ourselin, S.,
  Davidson, B., Hawkes, D., Clarkson, M.J.:
\newblock Hand--eye calibration for rigid laparoscopes using an invariant
  point.
\newblock International Journal of Computer Assisted Radiology and Surgery
  \textbf{11}(6) (2016)  1071--1080

\bibitem{Zhang1994}
Zhang, Z.:
\newblock Iterative point matching for registration of free-form curves and
  surfaces.
\newblock International journal of computer vision \textbf{13}(2) (1994)
  119--152

\bibitem{Du2010}
Du, S., Zheng, N., Xiong, L., Ying, S., Xue, J.:
\newblock Scaling iterative closest point algorithm for registration of m–d
  point sets.
\newblock Journal of Visual Communication and Image Representation
  \textbf{21}(5–6) (2010)  442 -- 452 Special issue on Multi-camera Imaging,
  Coding and Innovative Display.

\bibitem{Ventura2014}
Ventura, J., Arth, C., Reitmayr, G., Schmalstieg, D.:
\newblock A minimal solution to the generalized pose-and-scale problem.
\newblock In: The IEEE Conference on Computer Vision and Pattern Recognition
  (CVPR). (June 2014)

\bibitem{Sweeney2014}
Sweeney, C., Fragoso, V., H{\"o}llerer, T., Turk, M.:
\newblock gdls: A scalable solution to the generalized pose and scale problem.
\newblock In: Computer Vision--ECCV 2014.
\newblock Springer (2014)  16--31

\bibitem{Haralick1991}
Haralick, R.M., Lee, D., Ottenburg, K., Nolle, M.:
\newblock Analysis and solutions of the three point perspective pose estimation
  problem.
\newblock In: Computer Vision and Pattern Recognition, 1991. Proceedings CVPR
  '91., IEEE Computer Society Conference on. (Jun 1991)  592--598

\bibitem{Quan1999}
Quan, L., Lan, Z.:
\newblock Linear n-point camera pose determination.
\newblock IEEE Transactions on Pattern Analysis and Machine Intelligence
  \textbf{21}(8) (Aug 1999)  774--780

\bibitem{Lepetit2009}
Lepetit, V., Moreno-Noguer, F., Fua, P.:
\newblock Epnp: An accurate o (n) solution to the pnp problem.
\newblock International journal of computer vision \textbf{81}(2) (2009)
  155--166

\bibitem{Sweeney2015}
Sweeney, C., Kneip, L., Hollerer, T., Turk, M.:
\newblock Computing similarity transformations from only image correspondences.
\newblock In: The IEEE Conference on Computer Vision and Pattern Recognition
  (CVPR). (June 2015)

\bibitem{Zhang1991}
Zhang, Z., Faugeras, O.D.:
\newblock The first bmvc 1990 determining motion from 3d line segment matches:
  a comparative study.
\newblock Image and Vision Computing \textbf{9}(1) (1991)  10 -- 19

\bibitem{Kamgar2004}
Kamgar-Parsi, B., Kamgar-Parsi, B.:
\newblock Algorithms for matching 3d line sets.
\newblock IEEE Transactions on Pattern Analysis and Machine Intelligence
  \textbf{26}(5) (May 2004)  582--593

\bibitem{Bartoli2001}
Bartoli, A., Sturm, P.:
\newblock The 3d line motion matrix and alignment of line reconstructions.
\newblock In: Computer Vision and Pattern Recognition, 2001. CVPR 2001.
  Proceedings of the 2001 IEEE Computer Society Conference on. Volume~1. (2001)
   I--287--I--292 vol.1

\bibitem{Ramalingam2010}
Ramalingam, S., Taguchi, Y., Marks, T.K., Tuzel, O.:
\newblock P2$\pi$: A minimal solution for registration of 3d points to 3d
  planes.
\newblock In: Computer Vision--ECCV 2010.
\newblock Springer (2010)  436--449

\bibitem{Nister2004}
Nister, D.:
\newblock An efficient solution to the five-point relative pose problem.
\newblock IEEE Transactions on Pattern Analysis and Machine Intelligence
  \textbf{26}(6) (June 2004)  756--770

\bibitem{Stewenius2005}
Stewenius, H.:
\newblock Gr{\"o}bner Basis Methods for Minimal Problems in Computer Vision.
\newblock PhD thesis, Lund University (2005)

\bibitem{Stewenius2005b}
Stew{\'e}nius, H., Nist{\'e}r, D., Oskarsson, M., {\AA}str{\"o}m, K.:
\newblock Solutions to minimal generalized relative pose problems.
\newblock In: Workshop on omnidirectional vision. Volume~1. (2005) ~3

\bibitem{Kukelova2008}
Kukelova, Z., Bujnak, M., Pajdla, T.:
\newblock Automatic generator of minimal problem solvers.
\newblock In: Computer Vision--ECCV 2008.
\newblock Springer (2008)  302--315

\bibitem{Kukelova2012}
Kukelova, Z., Bujnak, M., Pajdla, T.:
\newblock Polynomial eigenvalue solutions to minimal problems in computer
  vision.
\newblock IEEE Transactions on Pattern Analysis and Machine Intelligence
  \textbf{34}(7) (July 2012)  1381--1393

\bibitem{Byrod2009}
Byr{\"o}d, M., Josephson, K., {\AA}str{\"o}m, K.:
\newblock Fast and stable polynomial equation solving and its application to
  computer vision.
\newblock International Journal of Computer Vision \textbf{84}(3) (2009)
  237--256

\bibitem{Marquardt1963}
Marquardt, D.W.:
\newblock An algorithm for least-squares estimation of nonlinear parameters.
\newblock Journal of the society for Industrial and Applied Mathematics
  \textbf{11}(2) (1963)  431--441

\bibitem{Fischler1981}
Fischler, M.A., Bolles, R.C.:
\newblock Random sample consensus: A paradigm for model fitting with
  applications to image analysis and automated cartography.
\newblock Commun. ACM \textbf{24}(6) (June 1981)  381--395

\end{thebibliography}
\end{document}